\documentclass[a4paper,conference]{IEEEtran}
\usepackage{times}
\usepackage{hyperref}

\usepackage{ifpdf}
\usepackage{booktabs}
\usepackage[table]{xcolor}

\usepackage{cite}
\usepackage{float}
\usepackage{tabulary}

\ifCLASSINFOpdf
\else
\fi
\usepackage{graphicx}

\usepackage{amsmath}
\usepackage{algorithmic}
\usepackage{array}
\usepackage{multirow}
\usepackage{threeparttable}

\ifCLASSOPTIONcompsoc
 \usepackage[caption=false,font=normalsize,labelfont=sf,textfont=sf]{subfig}
\else
 \usepackage[caption=false,font=footnotesize]{subfig}
\fi
\usepackage{fixltx2e}
\usepackage{stfloats}
\usepackage{url}
\begin{document}
%
% paper title
% Titles are generally capitalized except for words such as a, an, and, as,
% at, but, by, for, in, nor, of, on, or, the, to and up, which are usually
% not capitalized unless they are the first or last word of the title.
% Linebreaks \\ can be used within to get better formatting as desired.
% Do not put math or special symbols in the title.
\title{Image change detection with only a few samples}

\author{\IEEEauthorblockN{Ke Liu$^\ast$,
Zhaoyi Song$^\ast$\thanks{Ke Liu and Zhaoyi Song contributed equally to this work.} and
Haoyue Bai}
\IEEEauthorblockA{School of Software Engineering, Tongji University, Shanghai, China\\
\{juechen, vodkasoul, moonwhite\}@tongji.edu.cn}}

% use for special paper notices
%\IEEEspecialpapernotice{(Invited Paper)}

% make the title area
\maketitle

% As a general rule, do not put math, special symbols or citations
% in the abstract
\begin{abstract}
  This paper considers image change detection with only a small number of samples, which is a significant problem in terms of a few annotations available.
  A major impediment of image change detection task is the lack of large annotated datasets covering a wide variety of scenes. 
  Change detection models trained on insufficient datasets have shown poor generalization capability.
  To address the poor generalization issue, we propose using simple image processing methods for generating synthetic but informative datasets, and design an early fusion network based on object detection which could outperform the siamese neural network. Our key insight is 
  that the synthetic data enables the trained model to have good generalization ability for various scenarios. We compare the model 
  trained on the synthetic data with that on the real-world data captured
  from a challenging dataset, CDNet, using six different test sets. The results demonstrate that the synthetic 
  data is informative enough to achieve higher generalization ability than the insufficient real-world data. Besides, the experiment shows that utilizing a few (often tens of) samples to fine-tune 
  the model trained on the synthetic data will achieve excellent results.
\end{abstract}

% no keywords

% For peer review papers, you can put extra information on the cover
% page as needed:
% \ifCLASSOPTIONpeerreview
% \begin{center} \bfseries EDICS Category: 3-BBND \end{center}
% \fi
%
% For peerreview papers, this IEEEtran command inserts a page break and
% creates the second title. It will be ignored for other modes.
\IEEEpeerreviewmaketitle

\section{Introduction}
\begin{figure} 
    \centering
    \subfloat[Test images]{
\begin{minipage}[b]{0.30\linewidth}
\includegraphics[width=1\linewidth]{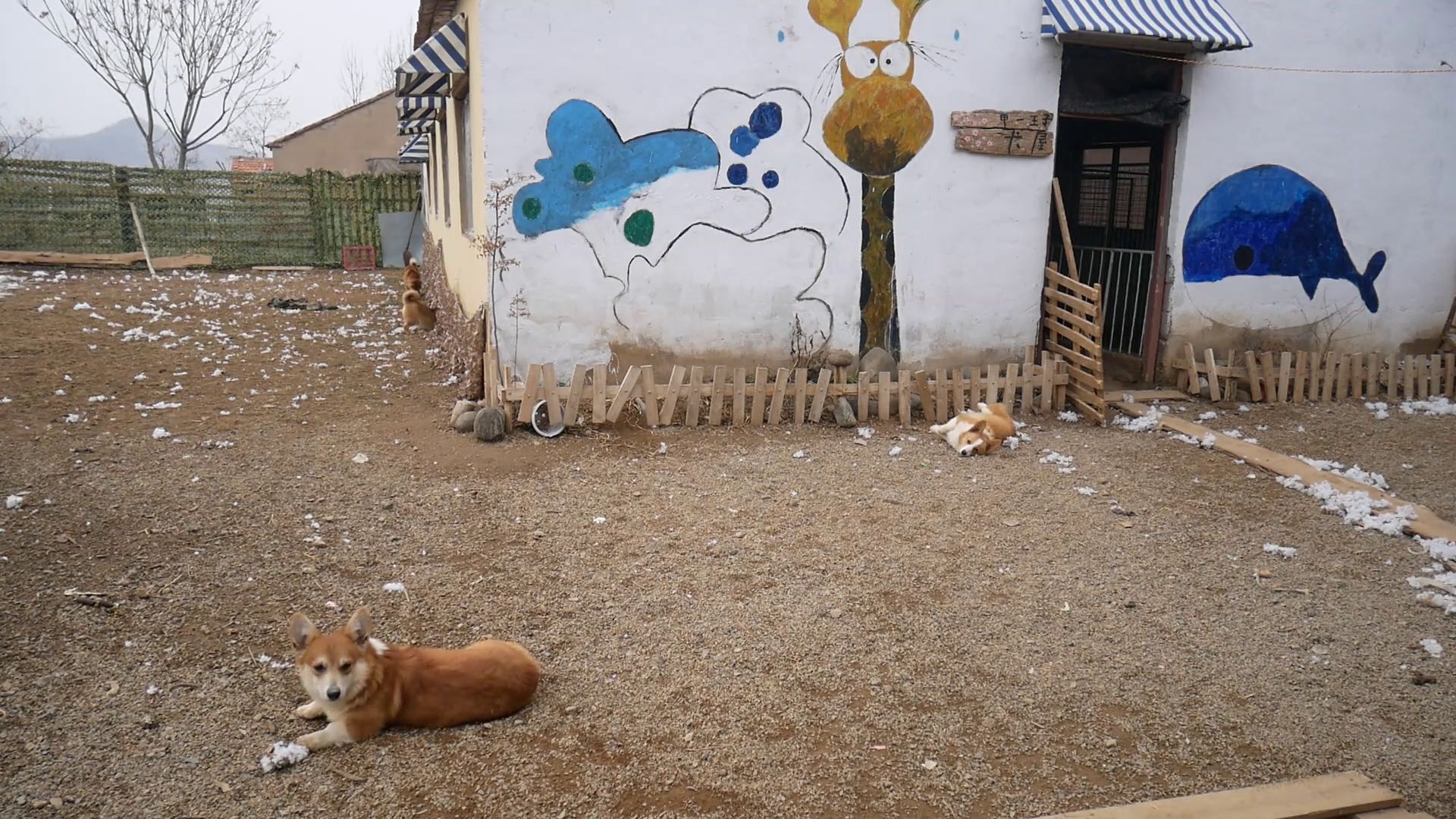}\vspace{4pt}
	   \includegraphics[width=1\linewidth]{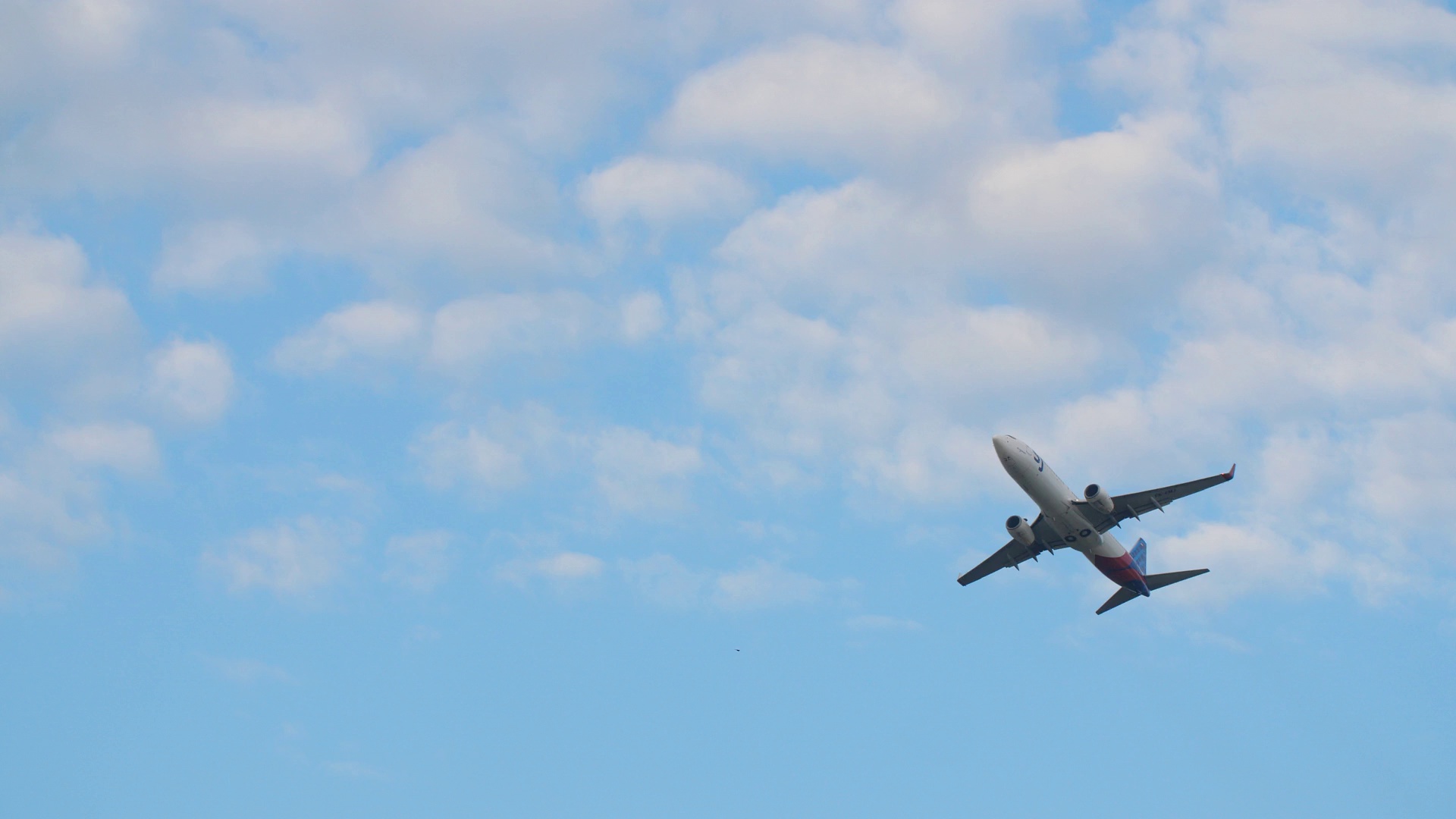}
\end{minipage}}
\label{i1}\hfill
    \subfloat[Reference images]{
\begin{minipage}[b]{0.30\linewidth}
\includegraphics[width=1\linewidth]{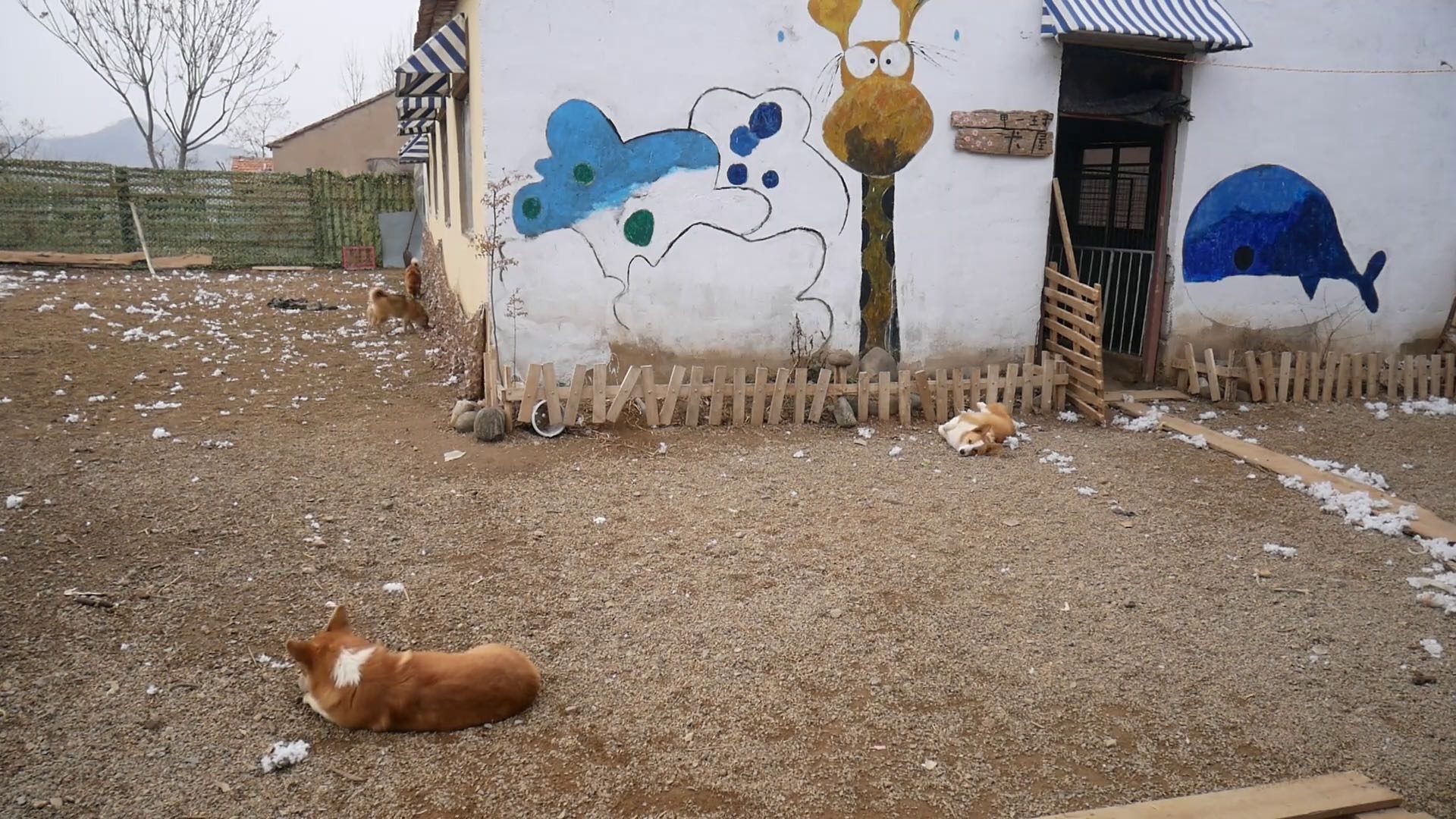}\vspace{4pt}
        \includegraphics[width=1\linewidth]{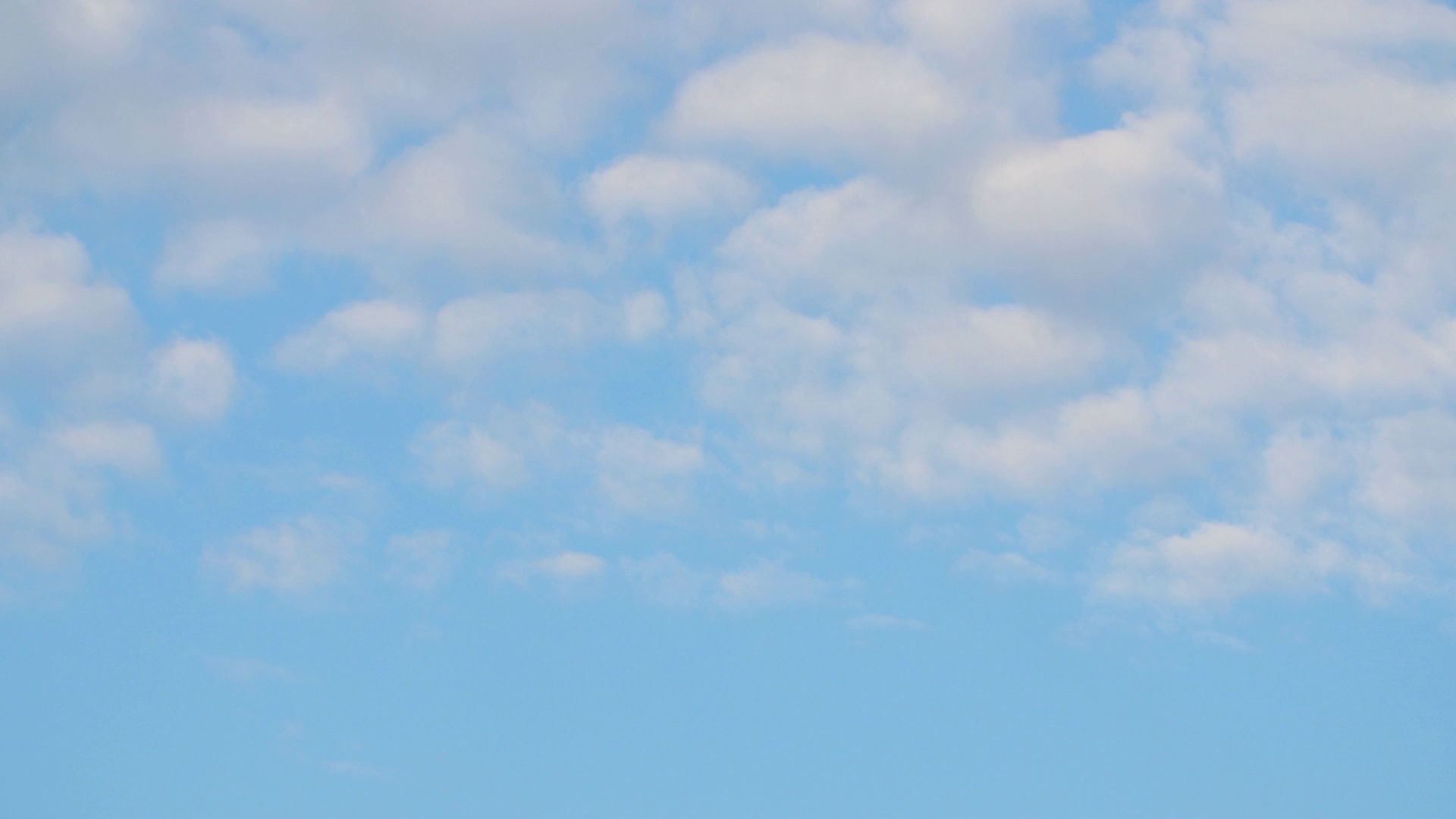}
\end{minipage}}
\label{r1}\hfill
    \subfloat[Detection Results]{
\begin{minipage}[b]{0.30\linewidth}
\includegraphics[width=1\linewidth]{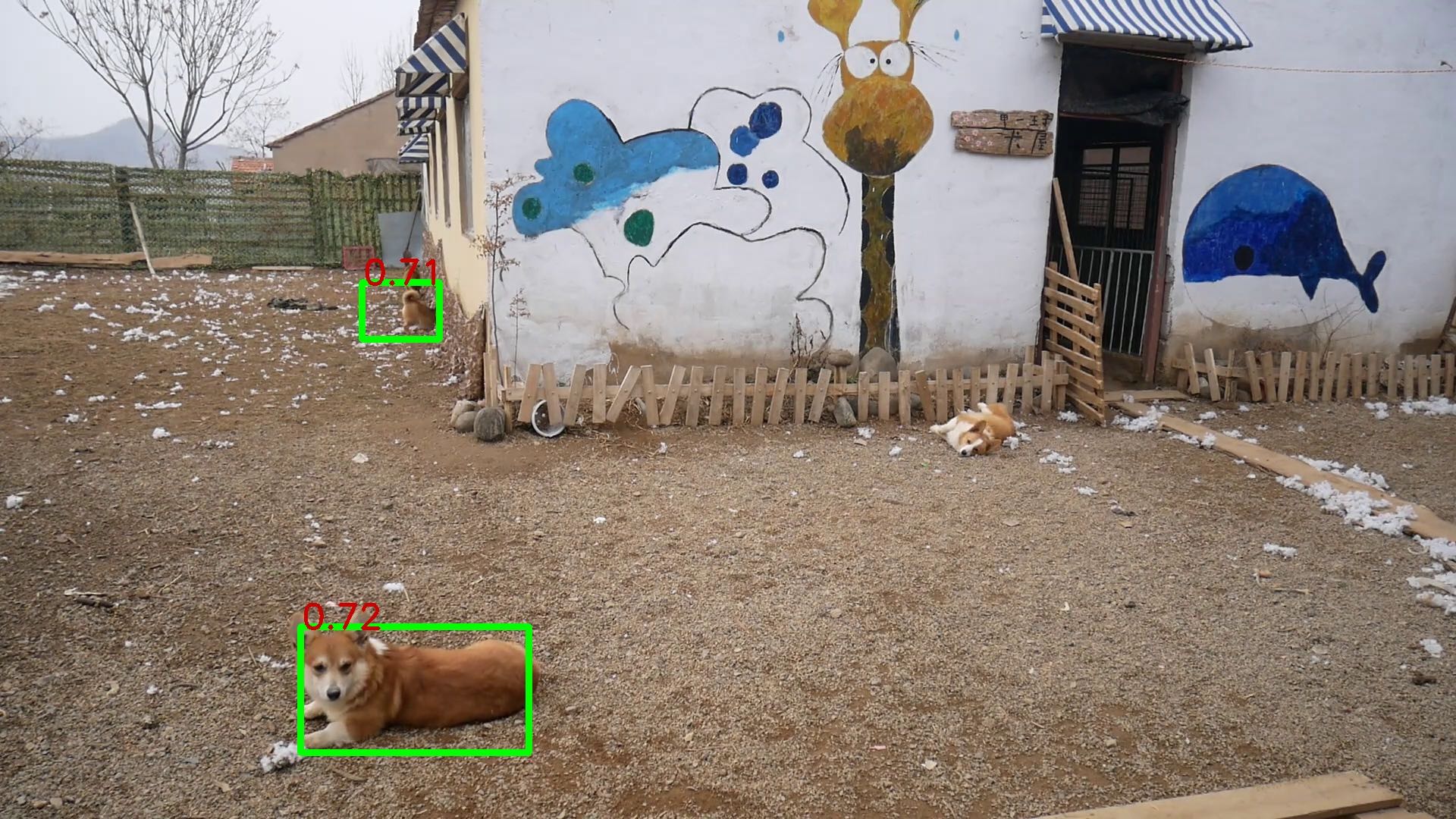}\vspace{4pt}
        \includegraphics[width=1\linewidth]{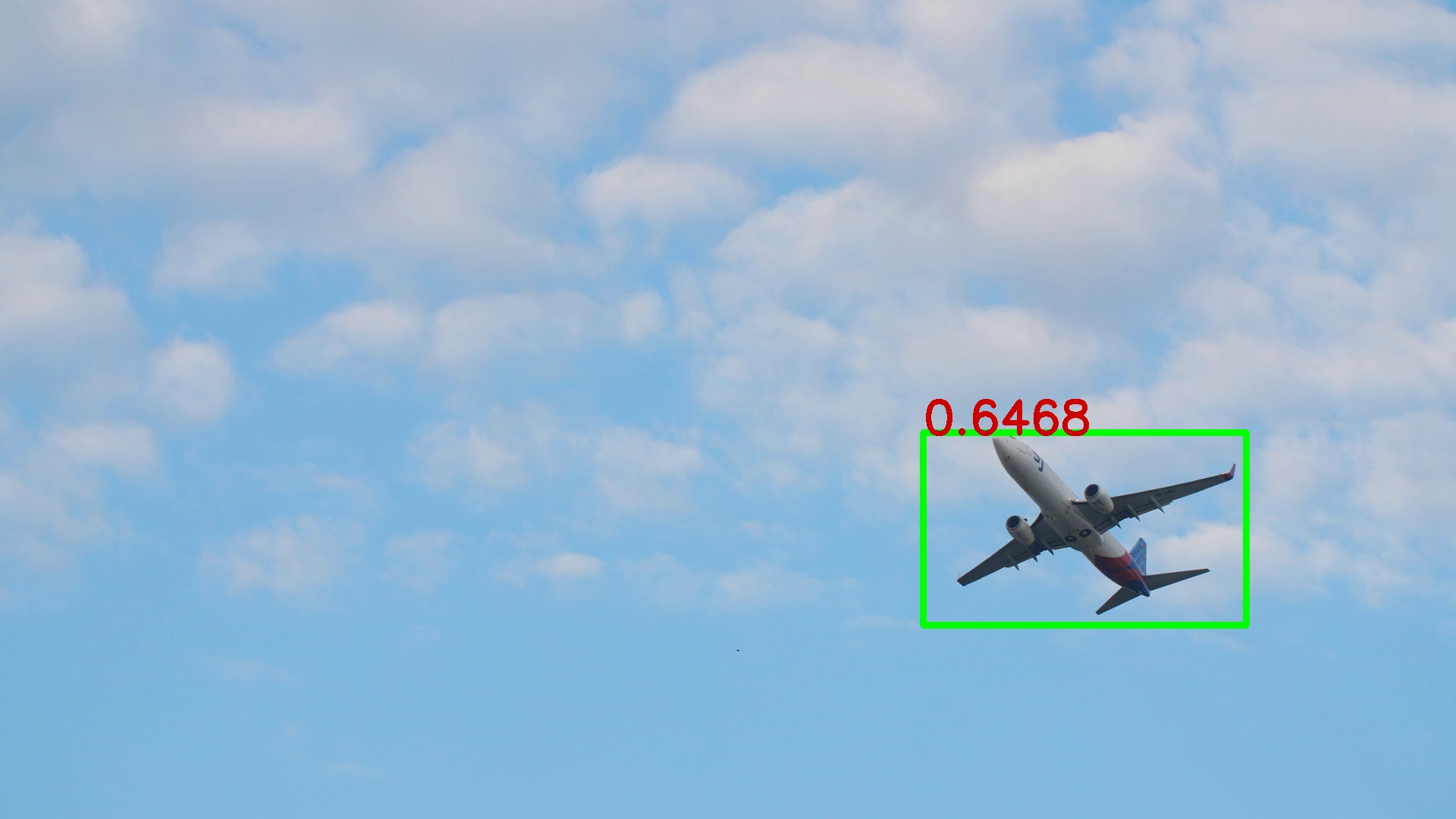}
\end{minipage}}
\label{d1}\hfill
    \caption{Some examples of the image pairs and detection results. 
    Images of dogs are on the top while images taken in campus are on the bottom. 
    Images (a) are the test images, on which we use boxes to mark the differences; 
    images (b) are the reference images; images (c) are the detection results including the bounding boxes and corresponding confidence. 
    }\label{fig:example} 
\end{figure}

This paper addresses the problem of change detection based on two images (i.e., the reference image and the test image), 
as illustrated in Figure~\ref{fig:example}.
% Imagine a situation where you take pictures at different times in the same place, and you wonder whether there is anything that has changed in the scene and, if changes do exist, in which part of the scene do the variations take place. 
Identifying changes in a region by comparing images taken at different times is called image change detection~\cite{radke2005image}, 
which is of practical significance and great research value.
% Generally speaking, automatic detection for such changes can be viewed as a visual change detection problem. 
For instance, it can detect and spot anomalies in an image compared with the reference image under normal conditions.
If the input is a series of frames derived from a video, 
then the algorithm can be applied to the consecutive two frames constantly to detect moving objects.
% A historical image (also referred to as reference image) at the specific location and a test image taken approximately at the same 
% place and from the same perspective can be used to identify the variations using computer vision techniques, such as semantic segmentation and object tracking. 

The history of change detection starts with the emergence of remote sensing~\cite{lu2004change}. 
% It has always been an interest for researchers in the field of computer vision ever since its birth. 
Nowadays, it is widely applied in video analysis and serves as a stepping stone for high-level scene understanding. 
Most of the traditional approaches are based upon image differencing and pixel-wise comparison. 
Many improving schemes, such as feature point extraction and image registration, 
are put forward to enhance the accuracy and efficiency of the algorithm. 
But not until the application of semantic segmentation was the study of change detection promoted to a new level, 
which signals a rapid transform from pixel-based methods to object-based approaches~\cite{chen2012object}. 
Recent studies have confirmed that the object-based change detection (OBCD) outperforms the pixel-based change detection (PBCD), 
especially when high spatial resolution imagery is used~\cite{blaschke2010object, huo2009fast}.

% However, despite the fantastic results the approaches mentioned above have achieved, 
However, the performance of OBCD is strongly related to the performance and accuracy of the classification algorithm~\cite{hussain2013change}, 
while the performance of current segmentation algorithms is highly dependent on the specified task~\cite{lu2015improving}.
Most of the change detection approaches based on segmentation are too `specific' since they tend to focus on 
the specific scenes (e.g., roads, campus). 
% When applying the models obtained from these approaches to a different scene, 
% we often find it a knotty problem since no satisfying outcome could be promised, 
% and we probably have to refine the model and adjust it according to the new requirements. 
It is prohibitively expensive to apply the overfitted models to new tasks or datasets.

Given the prevalence of neural networks, capturing and annotating training images has become a major concern. 
The lack of such images, which are difficult to acquire, can be a severe bottleneck in respect of the performance of a network.
There are lots of datasets available containing a few commonly occurring scenes and categories, 
but by no means are these datasets likely to scale to all possible scenes or categories.
Collecting massive annotations is a major impediment for change detection with strong generalization ability.

Considering the defects and deficiencies of these existing methods,
we are motivated to address the problem of change detection by using an automatically annotated synthetic dataset 
and a few manually annotated samples. 
Compared with other change detection methods, which often tend to depend on the image segmentation and classification techniques, 
our approach draws inspiration from object detection~\cite{zou2019object}, siamese neural networks~\cite{zhan2017change}, 
and transfer learning~\cite{pan2009survey}. 
More importantly, we provide a generalized synthetic dataset using digital image processing methods to tackle 
the problem of conflict between recognition accuracy and few annotations available.

% If we regard each difference as an object in the pair of images, then the problem becomes an object detection task.
% When adapting object detection methods to change detection problems, we can regard each change as an object in the image pairs.
Regarding each change as an object in the image pairs, we propose a novel method to resolve the change detection problem as an object detection task. Considering the model accepts two images as inputs and the task is to compare the image pairs by its very nature, 
we come up with two different network architectures: a 6-channel CNN using early fusion strategy as well as a siamese network.

In recent years, along with the advance of computer graphics and digital image processing,
making use of synthetically rendered scenes or objects has become a breach of the annotation barrier.
% However, in order to compete with the real-world datasets, much work remains to be done to make the synthetic images more realistic, 
% that is, bridging the ``reality gap''~\cite{tremblay2018training}. 
A large amount of such synthetic datasets are advanced, e.g., GTA V~\cite{richter2016playing}
, Sim4CV~\cite{muller2018sim4cv}, VIPER~\cite{richter2017playing}, to name a few. 
Extensive research has been carried out into generic image synthesis techniques designed for specific tasks, such 
as detection of vehicles and pedestrians~\cite{rozantsev2015rendering, marin2010learning}, 
% real-world scene understanding~\cite{handa2016understanding}, 
urban scene semantic segmentation~\cite{saleh2018effective}, or gesture 
recognition~\cite{athitsos2003estimating}.
However, to the best of our knowledge, there are still few if any synthetic datasets available for image change detection tasks.

To cope with the insufficiency of real-world data and enrich the feature information, 
we propose a simple but effective method to synthesize data for change detection. 
In order to simulate changes and differences between images, we randomly paste images on the background picture to form a new one. 
Thus the images we paste can be viewed as the `change' between the new picture and the original background picture. 
Crude as this synthesis method seems, it saves the efforts of the acquisition and manual annotation of images, 
which is rather tedious and laborious.

% More exactly, the focus is to make the pasted image not so abrupt.

Yet another tricky problem is how to make sure the pasted images fit in with the background picture. 
Encouragingly, the object detection methods care more about local region-based features for detection than the global scene layout~\cite{dwibedi2017cut}.
This means that the only stuff we should care about is the pasted image's fusion with the background, 
rather than the variations of image semantics caused by the pasted image.
In other words, our concern is not the realism of the synthetic images, but the local features we expect the neural network to learn.

Nonetheless, the synthetic data mean no exact equivalent of the real-world data since it is difficult to simulate the variable details in real-world scenes. 
Despite the convenience of generating synthetic data, models trained merely on such data have trouble generalizing to real-world scenarios. 
This is because the small variations in the images (e.g., shadows, illumination, contrast, and noises) have strong interference to the detection. 
% The model trained on the synthetic data  distinguish the small variations from the significant differences. 
% The small variations can be regarded as significant factors in some specific tasks, while in most cases, they should be ignored and dismissed.
The `change' is hard to define, since the `change' (such as shadows) labeled in a specific task may be dismissed in another scene or task.
Considering the variable definition of `change', which is largely dependent on the specific task, we refine the model with a small proportion of manually labeled data. 
% Regard the synthetic data as source domain and see the few samples as target domain, we fine-tune the model on it and think the model can learn the definition of `change' in the specific scenario. 
Regarding the synthetic data as the source domain and the new tasks as the target domain, we refine the model using a few samples from the target domain
% to exploit useful information from the source domain to enhance the performance on the target domain.
to learn the definition of the 'change'.
% Thus our goal is to train a model in different learning tasks so that some new learning tasks can be solved with few samples.
% Transfer learning (TL)  aims to enhance performance in a target domain by exploiting useful information from auxiliary or source domains
% when the labeled data in the target domain are insufficient or difficult to acquire~\cite{wu2017online}.
% It can be introduced to apply the information of the synthetic data (i.e., the source domain) and that
% of a few samples in the specific tasks (i.e., the target domain) to enhance generalization capability.
The results turn out that the fine-tuned model demonstrates a quite remarkable improvement.

In summary, the main contributions of this paper are listed as follows:

\begin{itemize}
  \item We propose an effective way to generate informative synthetic data after many trials. The model trained on the synthetic data outperforms that on the insufficient real-world data in various scenarios. Besides, to cope with the problem of minor changes, refining the model with a tiny proportion of real-world data works.
  % \item Inspired by~\cite{guo2018learning}, we propose a novel 6-channels network with the Contrast Measure Loss (CML). The key insight of CML is to maximize the distance 
  % of differences and minimize the distance of similarities. Experiment demonstrates that compared with the siamese network, the 6-channel network can take advantage of more information from the synthetic data and yield better performance. 
  \item Regarding ``change" between images as objects, we propose a novel early fusion network based on object detection. In comparison, a siamese network taking two images as input is also put forward. The architectures of them are shown in Figure~\ref{u_network}   
  Experiments demonstrate that compared with the siamese network, the early fusion network can take advantage of more information from the synthetic data and yield better detection performance. 
  \item To promote the research in the field of image change detection, we also provide nine manually annotated datasets, which are detailed in Section~\ref{sec:real}.
\end{itemize}

\section{Related Work}

Identifying the areas of difference between images is one of the most fundamental steps 
towards many common tasks. For instance, it is capable to check whether the photographic image of a book cover and 
its digital design are the same or different~\cite{wu2018spot}, or monitor the public infrastructure to 
check for encroachments~\cite{varghese2018changenet}.

As described in the previous section, the difference could either be the insertion or deletion of an 
object, or transformation of a particular object or event in the scene~\cite{rensink2002change}. 
One of the straightforward approaches to discovering and locating the dissimilarities is the 
pixel-wise comparison between a pair of images. Utilizing image registration, image noise suppression, 
and other technologies, the approach can get rid of the effects of noise and tiny variations~\cite{hussain2013change}. 
However, it often fails to obtain expected outcomes when confronting intense noise or significant 
variations~\cite{varghese2018changenet}.

Recently, there is a growing tendency to apply convolutional neural network in change detection tasks in view of its great fitting and generalization ability.
Furthermore, introducing siamese networks is an appealing alternative.
% Siamese neural networks have been widely applied in image retrieval~\cite{qi2016sketch}, object tracking~\cite{bertinetto2016fully}, signature verification~\cite{bromley1994signature}, among others. 
Due to its distinctive two tandem inputs and similarity measurement, 
the siamese network is competent to compare images in different contexts~\cite{zagoruyko2015learning}, particularly in the field of change detection. 
Zhan \emph{et al.}~\cite{zhan2017change} proposed a change 
detection method based on a siamese network using the weight contrastive loss for optical aerial images. 
Zhang \emph{et al.}~\cite{zhang2018change} used a siamese CNN for detecting topographic changes in the urban environment.

In the recent past, there has been quite a lot of benchmark datasets for change detection, among which 
video surveillance and remote sensing share the most common applications. In 2012, Goyette \emph{et al.}~\cite{goyette2012changedetection} 
introduced the changedetection.net (CDnet) benchmark, a video dataset devoted to the evaluation of change and 
motion approaches. Further in 2014, Wang \emph{et al.}~\cite{wang2014cdnet} expanded the dataset and evaluated up to 
14 change detection methods for the IEEE Change Detection Workshop 2014.

% Object detection is a fundamental step for automated image analysis in many computer vision applications. 
% Usually, it is performed by object detectors or background subtraction techniques~\cite{zhou2012moving}. 
% As mentioned in the previous section, if we regard each difference as an object in the pair of images, 
% then it becomes an object detection task. Region-based convolutional neural networks have played a significant 
% role in object detection, and CNN-based methods have shown leading results on detection benchmarks. An 
% outstanding incarnation, Faster R-CNN~\cite{ren2015faster}, achieves virtually real-time rates detection. Lately, 
% the anchor-free method has become popular research interests of object detection. 
% In 2019, Duan \emph{et al.}
% ~\cite{duan2019centernet} proposed a single-stage approach which utilizes the information of the center 
% point of an object to construct a triplet for object detection and demonstrated quite competitive performance.
% In 2019, Zhou \emph{et al.}~\cite{zhou2019objects} proposed a novel center point based approach, which presents an object by the centric 
% point of its bounding box to metamorphose the object detection problem into a standard keypoint estimation one,
% and demonstrated quite competitive performance.

% Considering the model we adopt accepts two images as inputs and the task is to compare the pair of images by its very nature, 
% it seems preferable to taking advantage of siamese neural networks.

% ============================================================
\section{Proposed Network}
To detect the difference between the image pair, we see the `change' as an object.
Then the detection task can be formulated as: 
\begin{equation}
    Y = F(X)    
\end{equation}
where image pair $(I_{test},I_{reference})$ serves as input $X$, and $Y$ contains the position of the `change' in the image. Here, we utilize the CNN-based network as the transformation function $F$.

\begin{figure*} 
    \centering
\subfloat[CUNet-EF]{\includegraphics[width=2.9in]{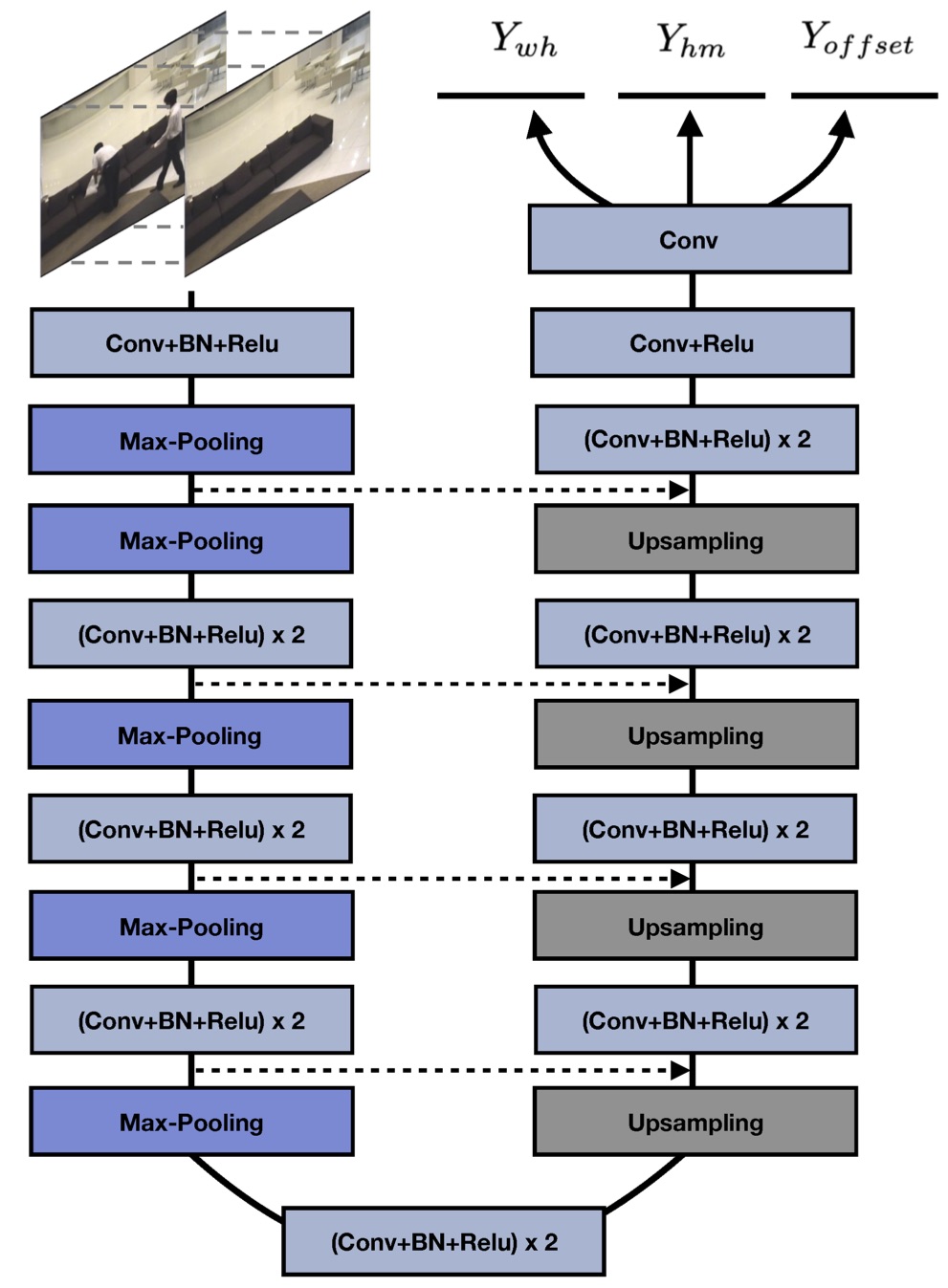} \label{fig_first_case}} 
\subfloat[CUNet-diff]{\includegraphics[width=3.5in]{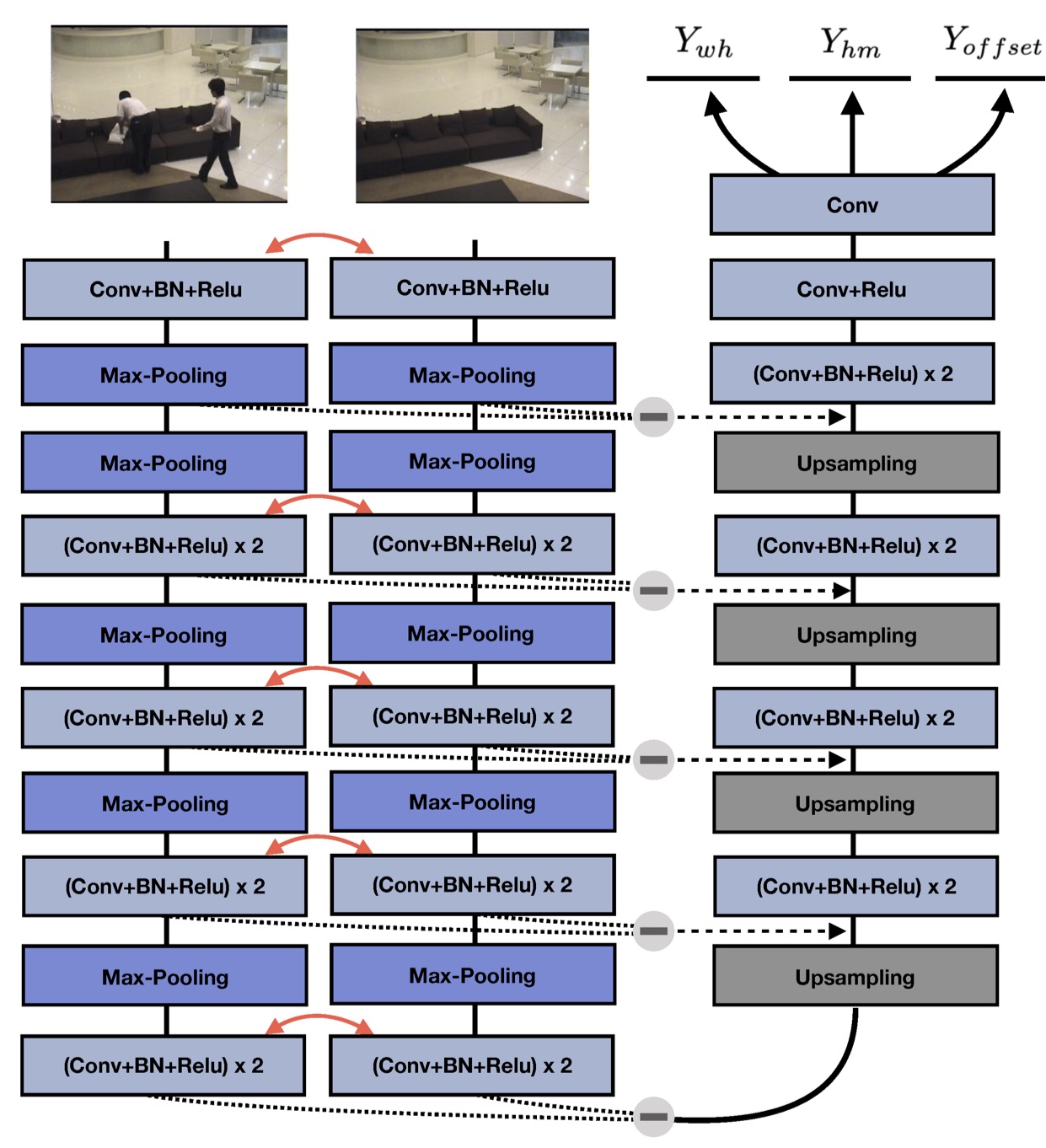} \label{fig_second_case}}
\caption{Details about the architectures. The left architecture is CUNet-EF and the right is CUNet-Diff. The red arrows mean the parameter weight is shared between two streams. The output of them are the map of hm, wh and offset which correspondingly indicates the center points of bounding boxes, the width and height of bounding boxes and the offset of the center points. }
\label{u_network} 
\end{figure*}

We propose two encoder-decoder network architectures: CUNet-EF and CUNet-Diff based on object detection. The two architectures are shown in Figure~\ref{u_network}.

\subsection{EF and Diff}
To detect the difference, the input of the network is an image pair. Therefore we need to change the usual CNN-based network to adapt it. The two adjustment schemes: Early Fusion (EF)  and Siamese-diff (Diff), have been adopted. EF means we stack the image pair and use it directly as the input of the network. While for the Diff, two streams as in a traditional Siamese network sharing the same architecture and weight serve as the encoding layers and the fusion operation is not done until the decoding part. Here, we fuse them by taking the absolute value of the difference between the encoders' outputs rather than stack them. \cite{daudt2018fully} shows that the two schemes have roughly the same performance in the field of image segmentation. We want to explore their performance in the object detection field and results are shown in Table~\ref{exp}.

\subsection{Encoder and Decoder}
The main backbone of our network is a basic encoder-decoder architecture U-Net~\cite{ronneberger2015u}, for it makes full use of location information and context information through downsampling, upsampling path and skip connections. However, such an encoder-decoder structure is usually used in the segmentation task and the final output size of it is usually the same as that of the decoding part to make full use of the feature information. Therefore, we make some changes so that it can work for object detection. Following the idea of \cite{zhou2019objects}, a map-based detection output of the same size as the decoding part is used.

\begin{figure}[!t] 
\centering 
\includegraphics[width=2.4in]{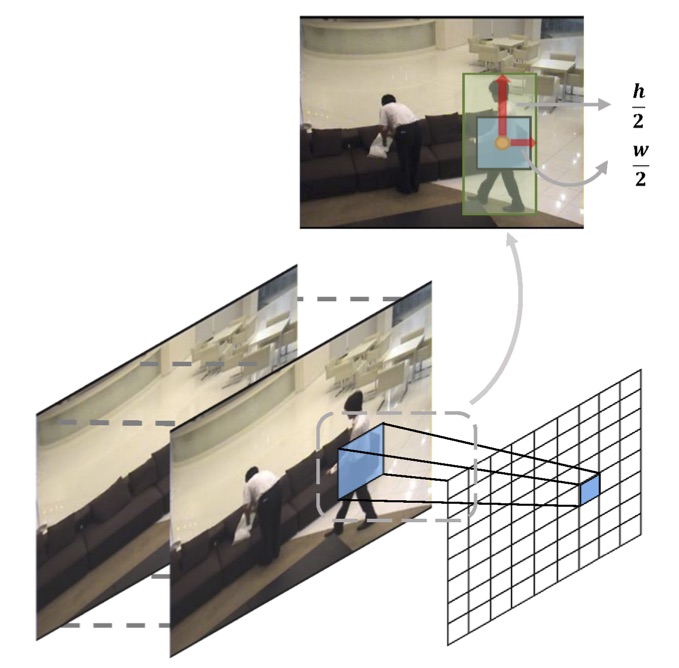}
\caption{Example of how the output maps work. An element on the map corresponds to an area on the original image. If the value of an element of $Y_{hm}$ is greater than the threshold (the blue block in the picture), it means the corresponding area on the original image has the center point of a `change'. By adding the position of this element with the value of the element of $Y_{offset}$ and mapping it to the original image, we get the final center point (the orange point). The element of $Y_{wh}$ indicates the width and height of the bounding box. The red rows in the picture display them. And the green box is the final result we get.} 
\label{detect} 
\end{figure}

The input resolution of the network is $512\times512$ and after pre-convolution, what fed into U-Net are the feature representations whose size is $128 \times 128$ instead of the original image. After the process of encoder and decoder, multiple convolution operations are followed to get the final outputs which including three maps $Y_{hm}$, $Y_{wh}$ and $Y_{offset}$. The size of the three maps is all $128 \times 128$ and an element of the $128 \times 128$ map can be viewed as a corresponding area of the $512 \times 512$ image. The element of $Y_{hm} \in [0,1]^{128 \times 128 \times 1}$ indicates whether there is a center point of the `change' in this area. $Y_{wh} \in R^{128 \times 128 \times 2}$ infers the width and height of bounding boxes of each center point. Due to the downsampling from 512 to 128, the integer division will lead to the loss of accuracy in the position of the center point. Hence, $Y_{offset}$ is needed to predict the offset of the center point in the original image. A direct description is shown in Figure~\ref{detect}.

\subsection{Loss}
Our loss function for an image pair consists of $L_{hm}$, $L_{wh}$ and $L_{offset}$.

For the object function about the center points, we use a variant of focal loss~\cite{lin2017focal}, according to \cite{zhou2019objects}. An ground truth center points map $G_{hm}\in[0,1]^{128 \times 128 \times 1}$ is generated using a Gaussian function, which means the true points are assigned 1 and the surrounding points of them are assigned value decreasing according to Gaussian function.
\begin{equation}
    G_{hm}^{x,y} = e^{-\frac{(x-\lfloor\frac{p_x}{4}\rfloor)^2+(y-\lfloor\frac{p_y}{4}\rfloor)^2}{2\sigma^2}}
\end{equation}
where p means the position of the center point in the original input and $\lfloor\frac{p}{4}\rfloor$ is the ground truth position mapped to the $G_{hm}$.
The major difference between our method and \cite{zhou2019objects} is that we only care about one object `change', so some complex processing is not necessary and is discarded.
The formulas of the $L_{hm}$ is
\begin{equation}
     L_{hm} = \frac{-1}{N}\sum_{x}\sum_{y}\left\{
             \begin{array}{lr}
             (1-Y_{hm}^{x,y})^\alpha \log(G_{hm}^{x,y}) & G_{hm}^{x,y}=1 \\
             (1-G_{hm}^{x,y})^\beta(Y_{hm}^{x,y})^\alpha\\ \quad \log(1-Y_{hm}^{x,y}) & otherwise\\
             \end{array}
\right.\\
\end{equation}
where N is the number of center points in the image pair($I_1$,$I_2$) and $\alpha$ and $\beta$ are hyper-parameters assigned 2 and 4. It means the further the distance of the prediction point away from the center point, the more severe the penalty will be.

For $L_{wh}$ and $L_{offset}$, we only take the true center points into consideration and use L1-loss to regress them.
\begin{equation}
    L_{offset} = \frac{1}{N}\sum_{p}|Y_{offset}^{p}-(\frac{p}{4}-\lfloor\frac{p}{4}\rfloor)|
\end{equation}
\begin{equation}
    L_{wh} = \frac{1}{N}\sum_{p}|Y_{wh}^{p} - G_{wh}^{p}|
\end{equation}
where $G_{wh}^{p}$ represents the ground truth of width and height of the center point $p$.

The overall loss is
\begin{equation}
    L = L_{hm} + \lambda_{wh}L_{wh} + \lambda_{offset}L_{offset}
\end{equation}
After multiple experiments, the balanced parameters $\lambda_{wh}=0.1$ and $\lambda_{offset}=1$ are taken in this task.

\section{Dataset Generation}
Due to the poor generalization performance of existing insufficient datasets about image change detection and the costly and time-consuming venture to collect and label the data, we are motivated to propose a simple but effective method to generate dataset.
Besides, we provide nine real-world datasets which are from different sources to evaluate the generalization capability of the synthetic data.

\subsection{Details about synthetic data}\label{restrictions}
\begin{figure} 
    \centering
    \subfloat[Instance Change]{
\begin{minipage}[b]{0.30\linewidth}
\includegraphics[width=1\linewidth]{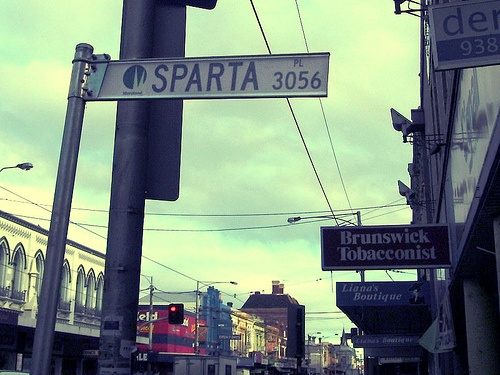}\vspace{4pt}
	   \includegraphics[width=1\linewidth]{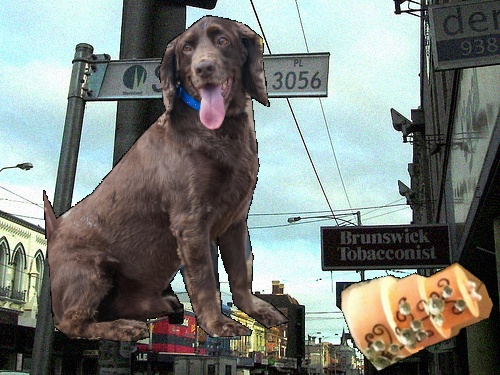}
\end{minipage}}
\label{transparent}\hfill
    \subfloat[Regular Cropped Change]{
\begin{minipage}[b]{0.30\linewidth}
\includegraphics[width=1\linewidth]{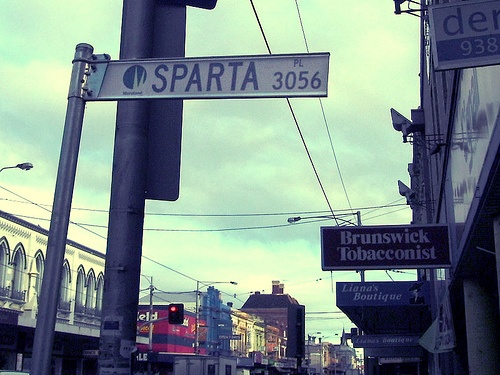}\vspace{4pt}
        \includegraphics[width=1\linewidth]{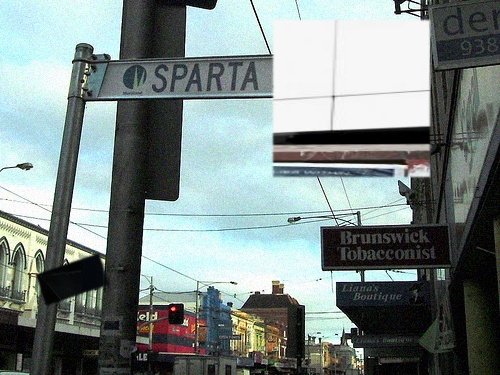}
\end{minipage}}
\label{cropped}\hfill
    \subfloat[Irregular Cropped Change]{
\begin{minipage}[b]{0.30\linewidth}
\includegraphics[width=1\linewidth]{images/exp5_i000049.jpg}\vspace{4pt}
        \includegraphics[width=1\linewidth]{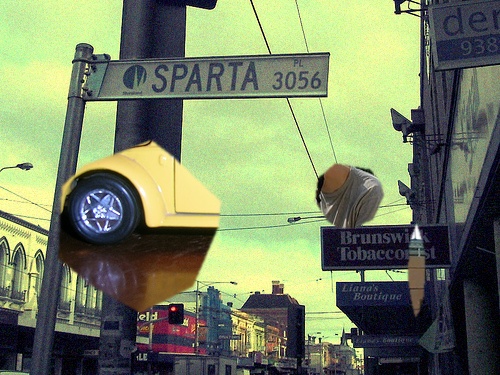}
\end{minipage}}
\label{1c}\hfill
	  \caption{Examples of the different `change'. Cropped Change is directly cropped from an image while the Instance Change has practical meaning. Besides, Multiple restrictions are also used.}
	  \label{example} 
\end{figure}

Here, we introduce the details about how to generate synthetic data. The process of generating an image pair with `change' can be summarized in the following steps. First, we choose a background image and some synthetic `change' needed to paste on. We attempt three different 'change', Regular Cropped Change, Instance Change and Irregular Cropped Change. Figure~\ref{example} displays examples of them. The second way is to randomly choose positions to paste them. It is noted that we do not care about the realism of the synthetic image pairs but care about the pasted parts' fusion with the background and the way to generate more local features. Therefore, some restrictions are imposed to make it more informative and natural. The details about the process are narrated in the following subsection, and the experiments about the performance of synthetic data are discussed in Section~\ref{result} 

\subsubsection{Details about `change'}
For Regular Cropped Change, the images we cut and paste are from the same dataset to ensure the image quality. Besides, we find the diversity of the size of `change' is significant. It is hard for a detector to predict the bounding box for a large `change' if it has not learned such information before. Therefore, we use the `Anchor box based' method to make more spatial information.

Anchor boxes, proposed in~\cite{ren2015faster}, are a set of predefined bounding boxes of a certain height and width to capture the scale and aspect ratio of the specific object. In the data generating view, we make use of it to hold the diversity of the aspect ratio and area of the `change'. After multiple experiments, the aspect ratios we adopted are $1:1,1:2,1:3,1:5,1:7$ with 50\% probability to rotate $90^{\circ}$ to turn the dominant character of width or height, and the corresponding ratio of generating numbers is about $3:3:3:2:2$. Towards the area, we define $S$ as the proportion of the area of `change' in the image, where $S_{small}$ refers to a range $[0.005,0.05)$, $S_{medium}$ refers to a range $[0.05,0.25)$ and $S_{large}$ refers to $[0.25,0.5)$. The corresponding proportion of them is about $4:2:1$.

\begin{figure}[!t] 
\centering 
\includegraphics[width=2.8in]{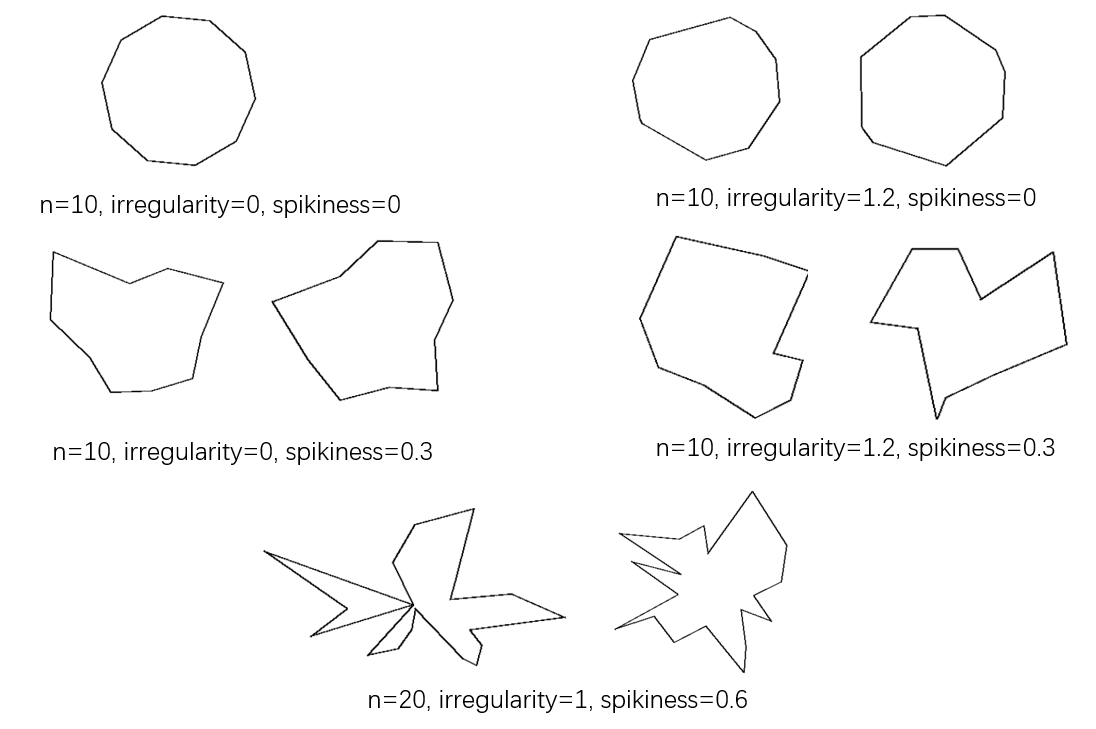}
\caption{An example of different shapes of Irregular Cropped Change when adopting different parameters.} 
\label{irregular} 
\end{figure}

Besides, to make the `change' looks more natural. In some cases, we crop a polygon instead of a rectangle from one image and paste it to the target image. This `change' is called Irregular Cropped Change. The basic idea about how to make it is to walk around the oval, taking a random angular step each time, and at each step put a point at a random radius. We use three parameters, n, irregularity and spikiness to control the shape of the polygon. Here n controls the number of edges of the polygon; irregularity controls whether or not the points are uniformly spaced angularly around the oval and spikiness controls how much the points vary from the oval. An example of it is shown in Figure~\ref{irregular}. Through a lot of trials, n is decided to be 10, the spikiness is randomly selected between 0 to 0.15, and the irregularity is randomly selected between 0.4 to 0.7.

For Instance change, the instance segmentation method is used to extract the instance. It is noted that the instances are from the same dataset with the background image. We think this method already ensures the diversity of `change' and do not need to use the `Anchor box based' method.
\subsubsection{Restrictions}
Rotation is applied to all the `change' to enrich the local feature information. A `change' with different rotation angles can reflect different feature information.

In addition, margin blur is applied to make the fusion more natural. To smooth the drastic change between the edge of the background and the `change', we apply gaussian blur to the edge. In our experiment, if the value is too high, the blur eﬀect is obvious but the results are not satisfying. So lastly, the value is randomly selected between 0.8 to 1.1 and makes the margin blur eﬀect not so easy to observe by eyes, but indeed makes the edges more natural from the perspective of pixel values. 

Furthermore, the presence of some noise in the image pairs is necessary to improve the robustness. Therefore, Gaussian noise and color change are randomly applied to an image of the image pair to make a disturbance.

\begin{figure}[!t] 
\centering 
\includegraphics[width=3in]{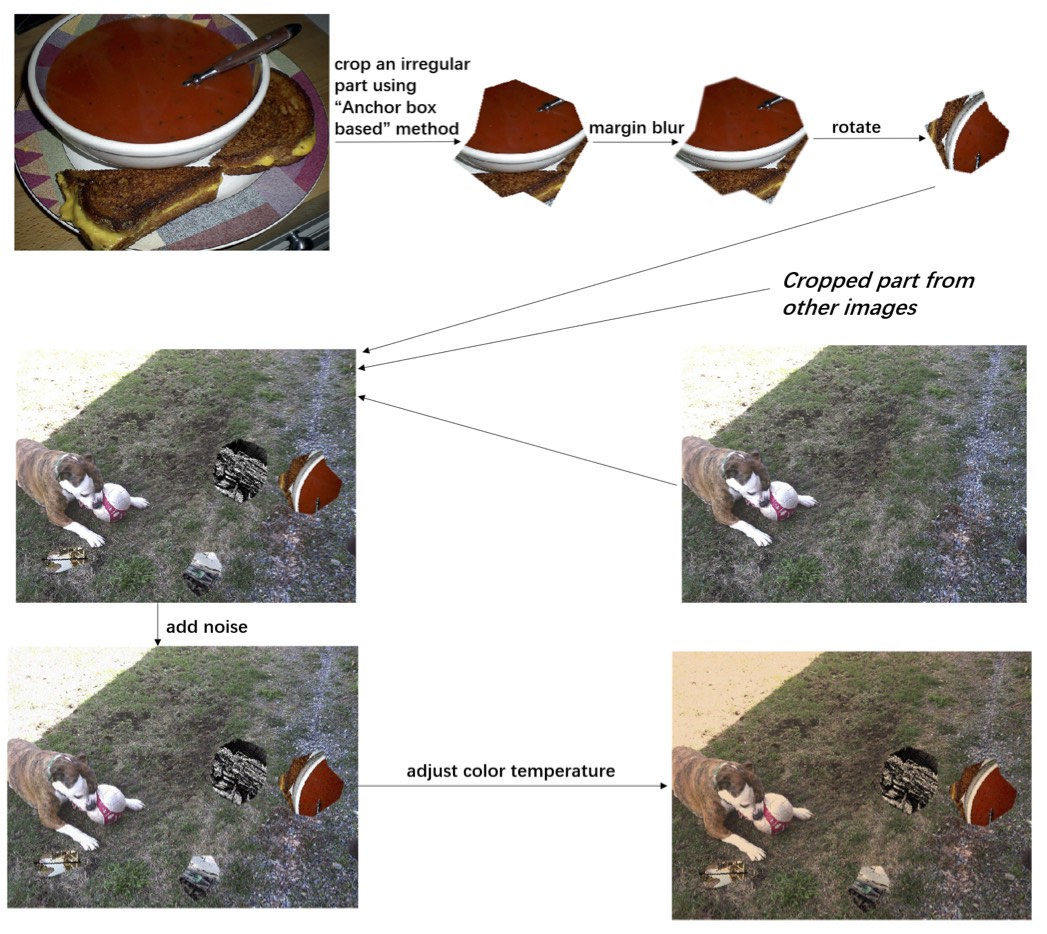}
\caption{An example of generating an image pair with Irregular Cropped Change and other restrictions.} 
\label{process_p} 
\end{figure}

An example of the process to generate an image pair with Irregular Cropped Change with restrictions is shown in Figure~\ref{process_p}

\subsection{Details about Real-world Datasets}\label{sec:real}
As mentioned before, the real-world dataset contains such objects is rare. But frames of some videos and reliable public datasets can be seen as image pairs that contain the `change'. CDNet~\cite{wang2014cdnet} is a challenging dataset for semantic change detection tasks, including frames of surveillance videos in various conditions. CCTV-Fights~\cite{perez2019detection} is a novel and challenging dataset containing 1,000 videos of real fights. SPID~\cite{wang2016spid} is a dataset collected for the pedestrian detection research works using surveillance images or videos. Some videos or images of them are sampled and manually labeled to adapt to image change detection in the object detection field. Besides, we collect some surveillance videos about pets from BiliBili, a video website. Furthermore, a dataset from a Chinese electricity company, which wants to detect the abnormal condition depending on spotting the `change' of devices, is also provided.  The details of the real dataset are shown in Table~\ref{real}. 

\begin{table}[!t] \renewcommand{\arraystretch}{1.3} \caption{Information about Real Data} \label{real} \centering \begin{tabular}{c|c|c} \hline \bfseries Name & \bfseries Number &\bfseries Sources \\ \hline Office & 97 & CDNet \\ 
 Skating & 78 & CDNet \\
 Bus Station & 55 & CDNet \\
 Sofa & 101 & CDNet \\
 Road & 407 & CDNet \\ 
 SJ-SPID & 260 & SPID \\ 
 Fight & 141 & CCTV-Fights \\ 
 Pets & 162 & Published videos in Bilibili \\ 
 Factory & 46 & A Chinese company \\ 
 \hline
\end{tabular} \end{table}
\section{Experiments and Results}
In this section, we evaluate two network architectures based on their performance and validate the effectiveness of the synthetic data. We have conducted two sets of experiments. In the first experiment, we want to find the best way to generate the synthetic data which have the best generalization capability. In the second set of experiments, we want to explore whether the different definition of `change' is a significant obstacle and whether utilize the few samples can overcome it. 

\subsection{Implementation Details}
\subsubsection{Generating}
When generating, the COCO~\cite{lin2014microsoft} dataset serves as the image source. An image from COCO is selected as the primary image, and `change' objects up to 5 are pasted to it to form an image pair. Each Cropped Change is randomly cropped from a COCO image. And for Instance Change, it is randomly selected from an instance set, which is produced by doing instance segmentation on COCO.

\begin{table}[] \renewcommand{\arraystretch}{1.3} \caption{Generating Strategies} \label{train} \centering \begin{tabulary}{\columnwidth}{c|c|C} \hline \bfseries Name & \bfseries Number &\bfseries Characteristics \\ \hline
 Exp.1 & 4780 & Cropped Change with margin blur  \\ 
 Exp.2 & 4780 & Instance Change with rotation and margin blur\\ 
 Exp.3 & 4780 & Cropped Change with rotation and margin blur\\
 Exp.4 & 4780 & Irregular Cropped Change with rotation and margin blur \\
 Exp.5 & 4780 & Blend of Exp.2 and Exp.3 with ratio 1:1 \\
 Exp.6 & 4780 & Blend of Exp.2 and Exp.4 with ratio 1:1 \\ 
 Exp.7 & 4780 & Blend of Exp.3 and Exp.4 with ratio 1:1 \\ 
 Exp.8 & 4779 & Blend of Exp.2, Exp.3 and Exp.4 with ratio 1:1:1 \\ 
 Exp.9 & 234 & Real Dataset: blend of Skating, Bus Station and Sofa 
 \\ \hline
 \end{tabulary} 
\end{table}

 \subsubsection{Training}
The data augmentation methods such as random flip, random scaling, cropping and color jittering are adopted to improve the performance. We use Adam to optimize this task and the initial learning rate is $1e^{-4}$. All of the networks are computed on a single GTX 2080 Ti.
 
 \subsubsection{Evaluation}To measure the performance of the model, we adopt Average Precision (AP) at IOU of 0.5 in all our experiments.

 \subsection{Details about experiments and results}\label{result}

 \subsubsection{Performance of synthetic data and network architecture}
 To explore which `change object' with restriction can generate excellent synthetic data, we attempt  different composition strategies and  generate eight different datasets. They all serve as training datasets. Besides, a real-world dataset consisting of Skating, BusStation and Sofa are utilized to represent the insufficient real-world data to compare the performance with synthetic datasets. The details about them are listed in Table~\ref{train}. Besides, six real datasets, Road, Office, Factory, Fight, Pets, SJ-SPID  serve as the testsets to validate the performance. When training, batch-size 32 is taken and the epoch is 240 with learning rate drop 10$\times$ every 80 epochs.
 
 The results are shown in Tabel~\ref{exp}. As a synthetic dataset can not perform the best result on all the testsets, we use the Euclid distance to measure the generalization capability of them. It can be formulated as:
 \begin{equation}
     Distance = ||X-X_{best}||_2
 \end{equation}
 where $X$ represents the set consisting of the test results of a network trained from a training dataset and $X_{best}$ is the set consisting of the best test result in each testset.
 
The results demonstrate that the insufficient real-world data can not overcome the gap between different `change' tasks and have a poor performance in the tastsets while all the synthetic datasets which possess imitation feature information outperform it.
 
 As for the comparison among synthetic datasets, it is obvious that the Instance Change is the worst one. This is because there are inevitable losses in the process of instance segmentation and the appearance of some instances extracted is not satisfactory. Furthermore, the blend of different objects may cause an improvement. Besides, the results also show that different architectures have different preferences for the synthetic data. The best performance of CUNet-Diff is on Exp.6 while CUNet-EF is on Exp.7.
 
 When comparing network architectures, although in\cite{daudt2018fully}, the two architectures are so well-matched in the image semantic segmentation field about change detection, in the object detection field, the results indicate CUNet-EF is more suitable for this task.
 
 Therefore, we think CUNet-EF is better than CUNet-Diff  and the blend of `Cropped Change with rotation and margin blur' and 'Irregular Cropped Change with rotation and margin blur' (i.e. Exp.7) is the best way to generate data.

 Although the results show that synthetic data contributes to a drastic improvement when compared with the insufficient real-world data and performs the promising generalization capability, the best results of SJ-SPID, Fight and Pets are not ideal. We think what leads to it is what is `change'. The meaning of it is that the `change' on which different scenarios focus is different. For a change in the image pair, in some scenarios, it is taken into consideration but in the other scenarios, it is viewed as noise. We believe a few samples from the scenario contain the definition of `change' and make use of them can have an improvement.  Experiments with such samples follow to demonstrate it.
 
 \begin{table*}[!t] \renewcommand{\arraystretch}{1.3} \caption{Results of different synthetic datasets on six real-world datasets} \label{exp} \centering
 \begin{threeparttable}
 \begin{tabular}{c c c c c c c c c c c c} \hline 
 \bfseries &  \bfseries & \bfseries Exp.1 &\bfseries Exp.2 &\bfseries Exp.3 &\bfseries Exp.4 &\bfseries Exp.5 &\bfseries Exp.6 &\bfseries Exp.7 &\bfseries Exp.8 &\bfseries Exp.9  \\ \hline
 \multirow{2}*{Office} &CUNet-Diff&75.7\%&73.9\%&97.5\%&92.7\%&81.8\%&96.4\%&\cellcolor{lightgray}$\mathbf{100\%}$&83.9\% &73.8\%\\ 
 &CUNet-EF&13.5\%&\cellcolor{lightgray}$\mathbf{100.0\%}$&90.0\%&95.9\%&\cellcolor{lightgray}$\mathbf{100\%}$\%&\cellcolor{lightgray}$\mathbf{100\%}$&97.1\%&\cellcolor{lightgray}$\mathbf{100\%}$&94.7\%\\\hline
 \multirow{2}*{Road} &CUNet-Diff&72.6\%&45.3\%&54.8\%&62.3\%&42.1\%&64.2\%&63.0\%&42.1\%&0.5\%\\
 &CUNet-EF&83.9\%&58.6\%&84.0\%&\cellcolor{lightgray}$\mathbf{86.5\%}$&76.4\%&82.2\%&85.0\%&76.7\%&0.4\%\\\hline
 \multirow{2}*{Factory} &CUNet-Diff&55.8\%&49.9\%&50.7\%&60.3\%&61.3\%&63.8\%&55.3\%&50.9\%&2.5\%\\
 &CUNet-EF&68.5\%&45.7\%&\cellcolor{lightgray}$\mathbf{75.0\%}$&61.7\%&65.0\%&58.6\%&74.2\%&60.4\%&5.2\%\\\hline
 \multirow{2}*{SJ-SPID} &CUNet-Diff&51.7\%&54.9\%&46.6\%&43.0\%&49.1\%&55.7\%&48.2\%&43.8\%&2.1\%\\
 &CUNet-EF&47.6\%&50.9\%&68.7\%&60.1\%&54.1\%&54.0\%&\cellcolor{lightgray}$\mathbf{71.5\%}$&56.3\%&22.9\%\\\hline
 \multirow{2}*{Fight} &CUNet-Diff&38.9\%&25.1\%&26.3\%&32.7\%&25.0\%&31.6\%&32.3\%&27.8\%&9.7\%\\
 &CUNet-EF&40.3\%&32.5\%&54.0\%&46.7\%&39.9\%&34.4\%&\cellcolor{lightgray}$\mathbf{58.1\%}$&34.6\%&22.9\%\\\hline
 \multirow{2}*{Pets} &CUNet-Diff&26.8\%&33.3\%&40.3\%&45.3\%&36.2\%&\cellcolor{lightgray}$\mathbf{47.8\%}$&45.9\%&46.0\%&1.9\%\\
 &CUNet-EF&25.2\%&27.5\%&32.5\%&27.9\%&32.1\%&29.6\%&35.7\%&24.0\%&1.6\%\\\hline
  \multirow{2}*{Distance\tnote{*}} &CUNet-Diff&0.485&0.677&0.574&0.481&0.650&0.398&0.464&0.671&1.503\\
 &CUNet-EF&0.945&0.592&0.190&0.291&0.329&0.386&\cellcolor{lightgray}$\mathbf{0.126}$&0.407&1.343\\\hline

 \hline
\end{tabular} 
\begin{tablenotes}
        \footnotesize
        \item[*] Distance measures the generalization capability and the lowest one is the best one.
      \end{tablenotes}
\end{threeparttable}
\end{table*}

\begin{figure*}[!t]
    \centering
    \subfloat[Test images]{
\begin{minipage}[b]{0.12\linewidth}
\includegraphics[width=1\linewidth]{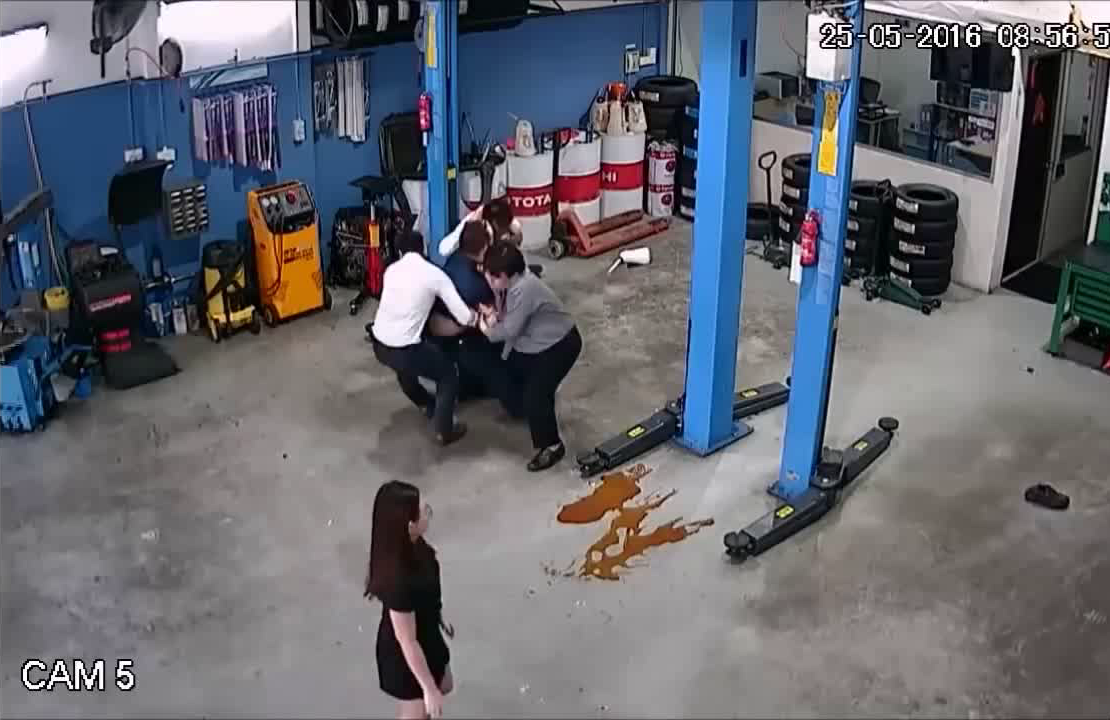}\vspace{4pt}
	   \includegraphics[width=1\linewidth]{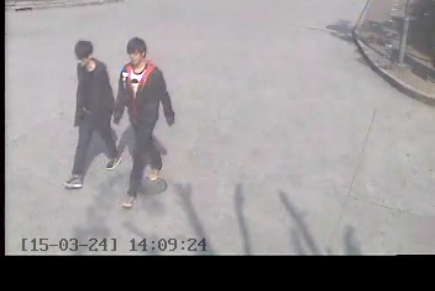}\vspace{4pt}
	   \includegraphics[width=1\linewidth]{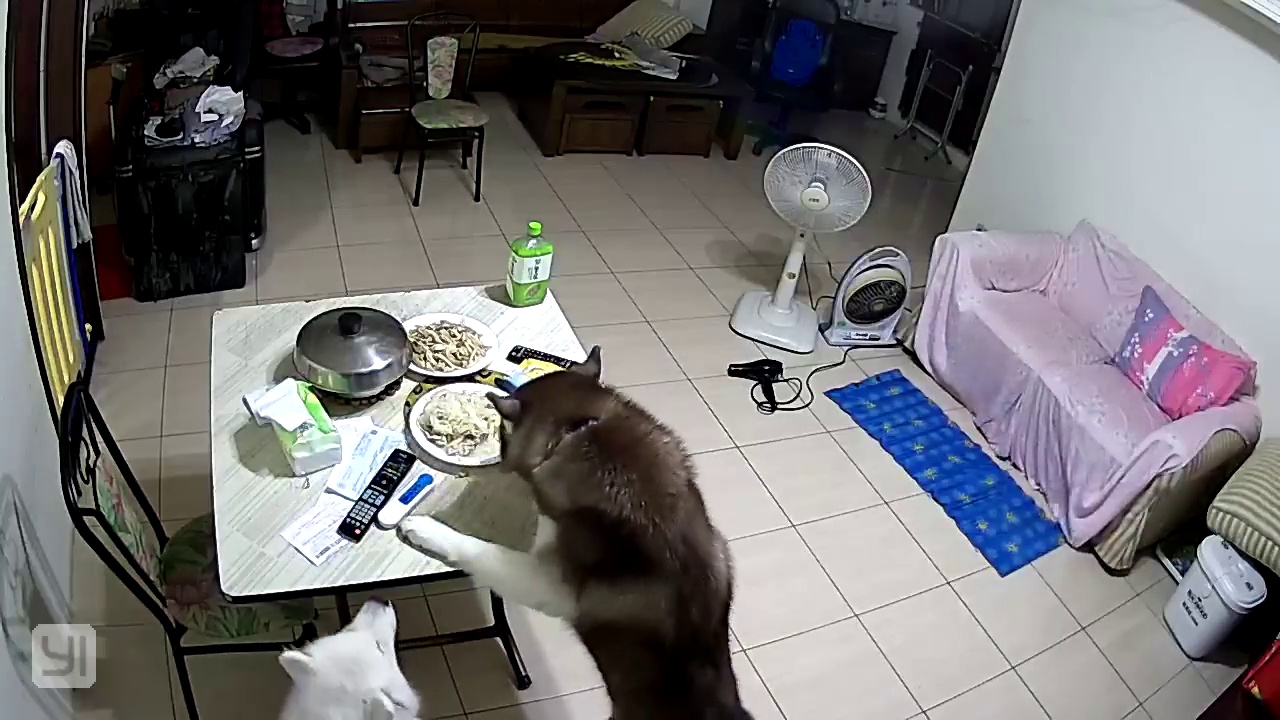}
\end{minipage}}
\label{transparent}\hfill
    \subfloat[Reference images]{
\begin{minipage}[b]{0.12\linewidth}
\includegraphics[width=1\linewidth]{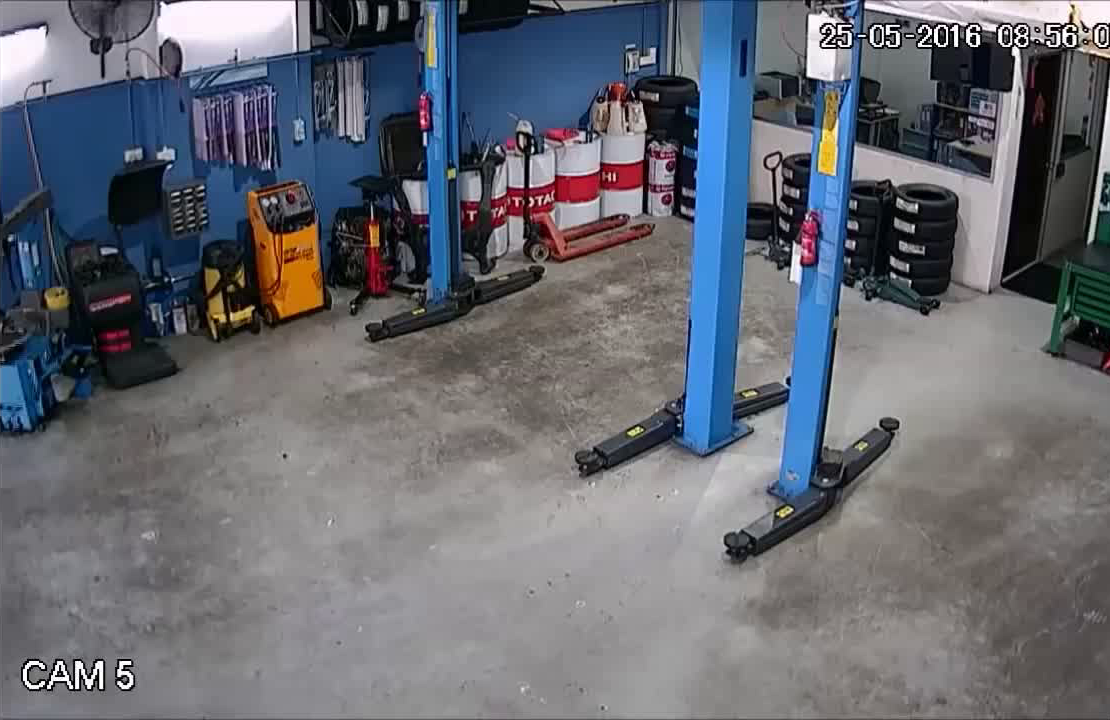}\vspace{4pt}
	   \includegraphics[width=1\linewidth]{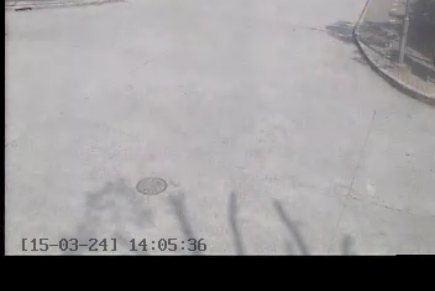}\vspace{4pt}
	   \includegraphics[width=1\linewidth]{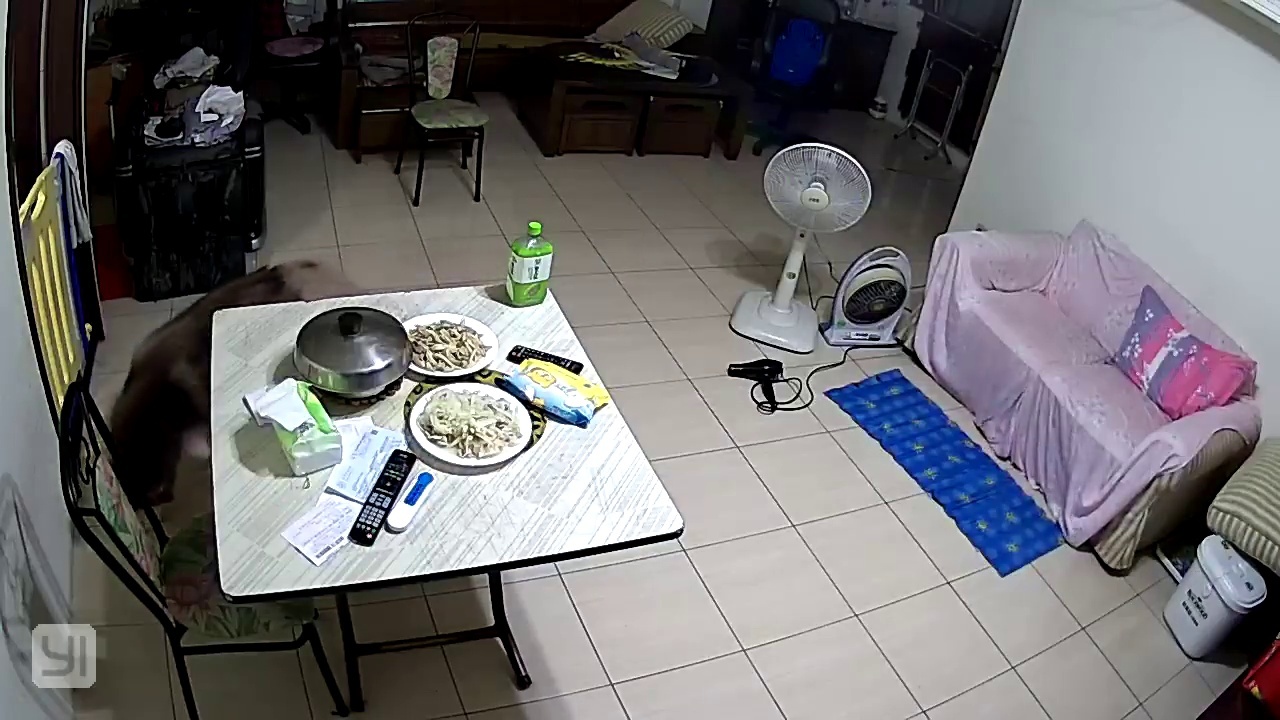}
\end{minipage}}
\label{4a}\hfill
    \subfloat[Ground Truth]{
\begin{minipage}[b]{0.12\linewidth}
\includegraphics[width=1\linewidth]{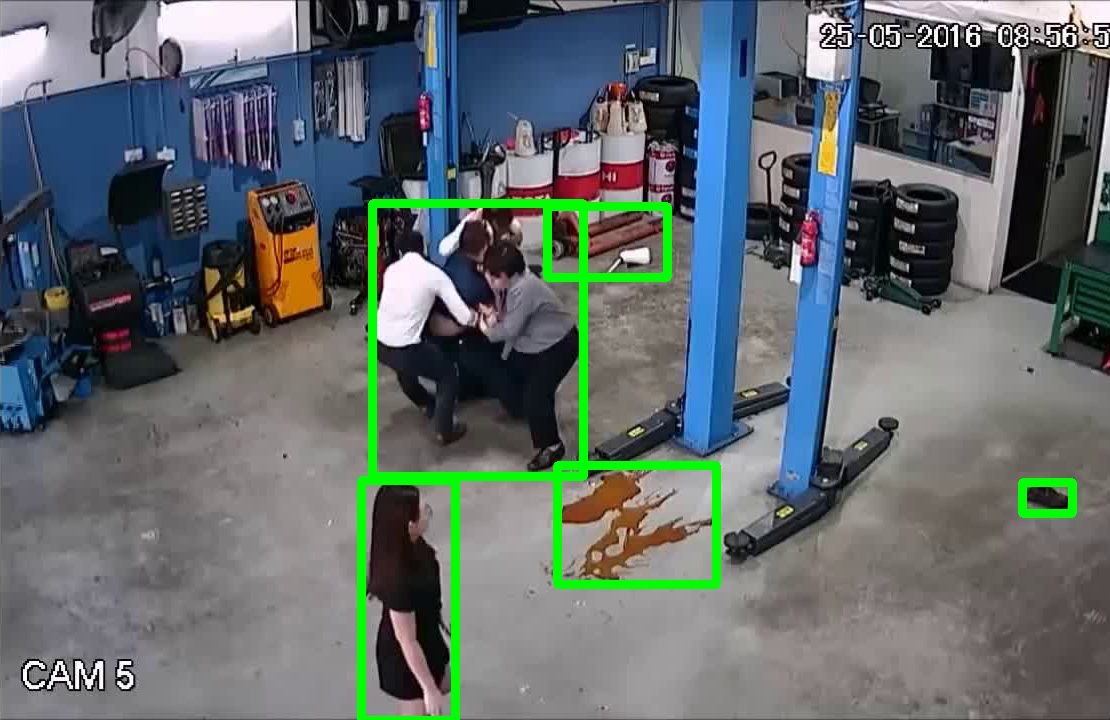}\vspace{4pt}
	   \includegraphics[width=1\linewidth]{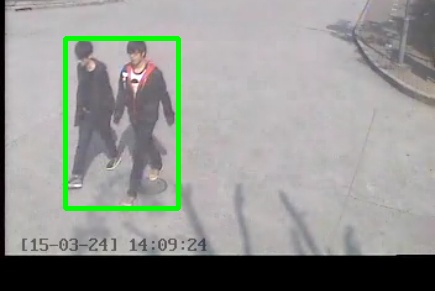}\vspace{4pt}
	   \includegraphics[width=1\linewidth]{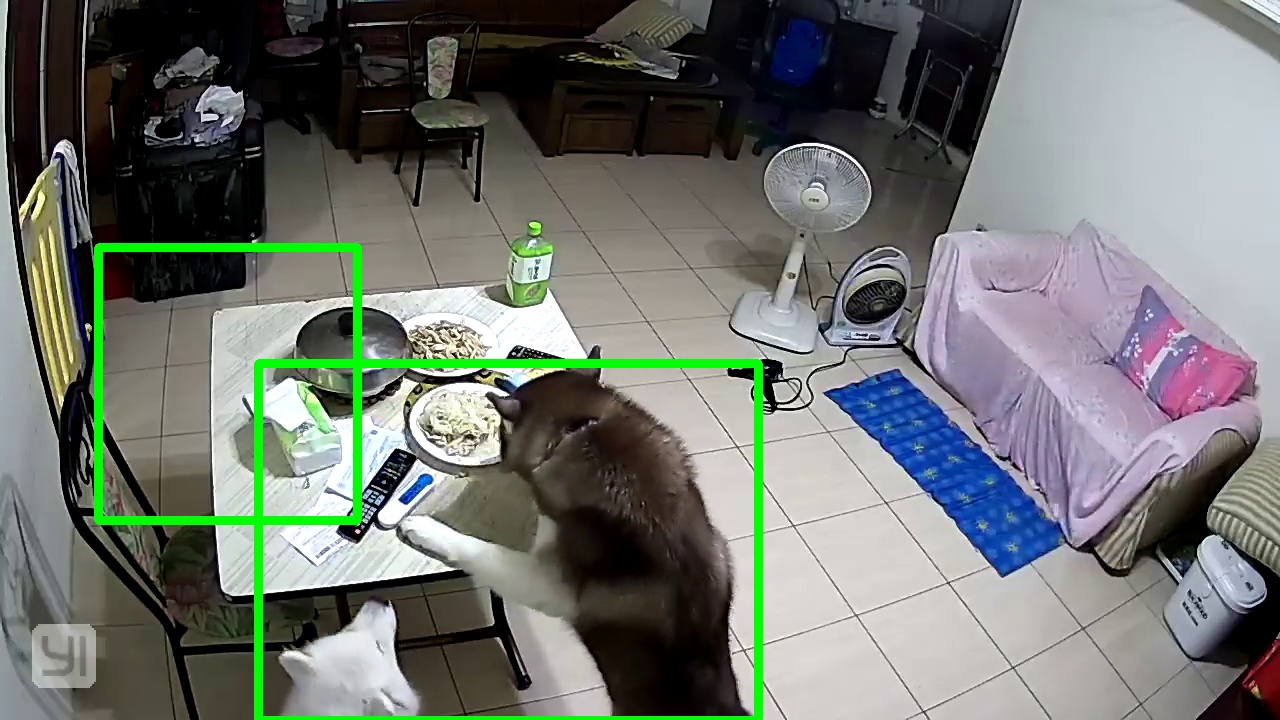}
\end{minipage}}
\label{4b}\hfill
    \subfloat[Without Fine-tuning]{
\begin{minipage}[b]{0.12\linewidth}
\includegraphics[width=1\linewidth]{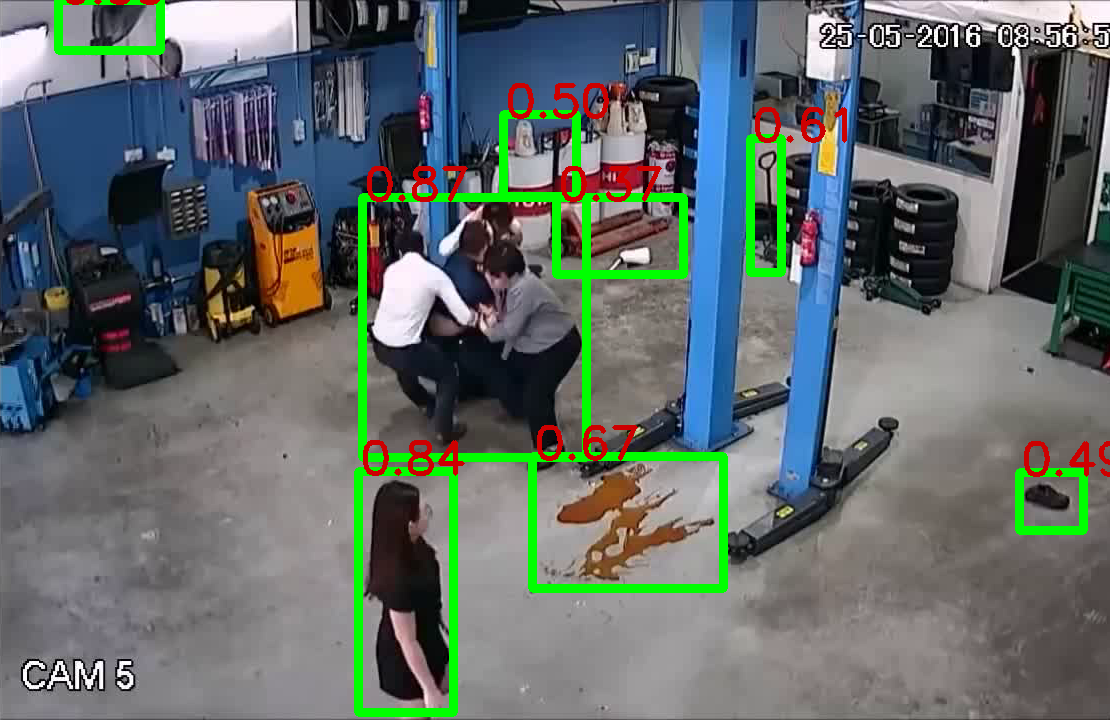}\vspace{4pt}
	   \includegraphics[width=1\linewidth]{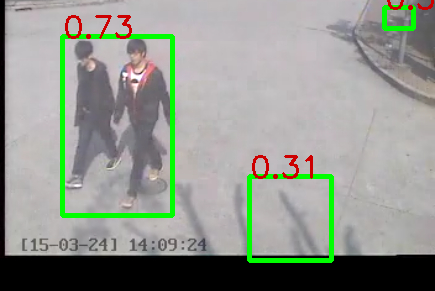}\vspace{4pt}
	   \includegraphics[width=1\linewidth]{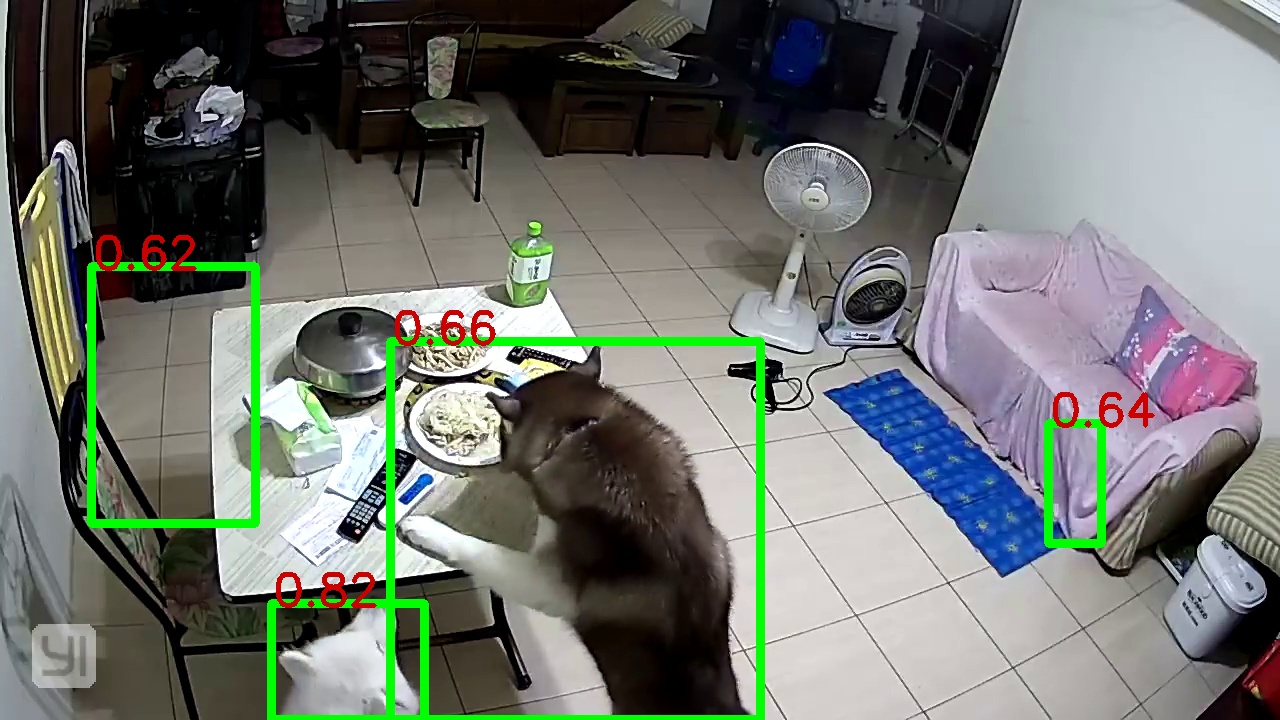}
\end{minipage}}
\label{4c}\hfill
   \subfloat[Mixture]{
\begin{minipage}[b]{0.12\linewidth}
\includegraphics[width=1\linewidth]{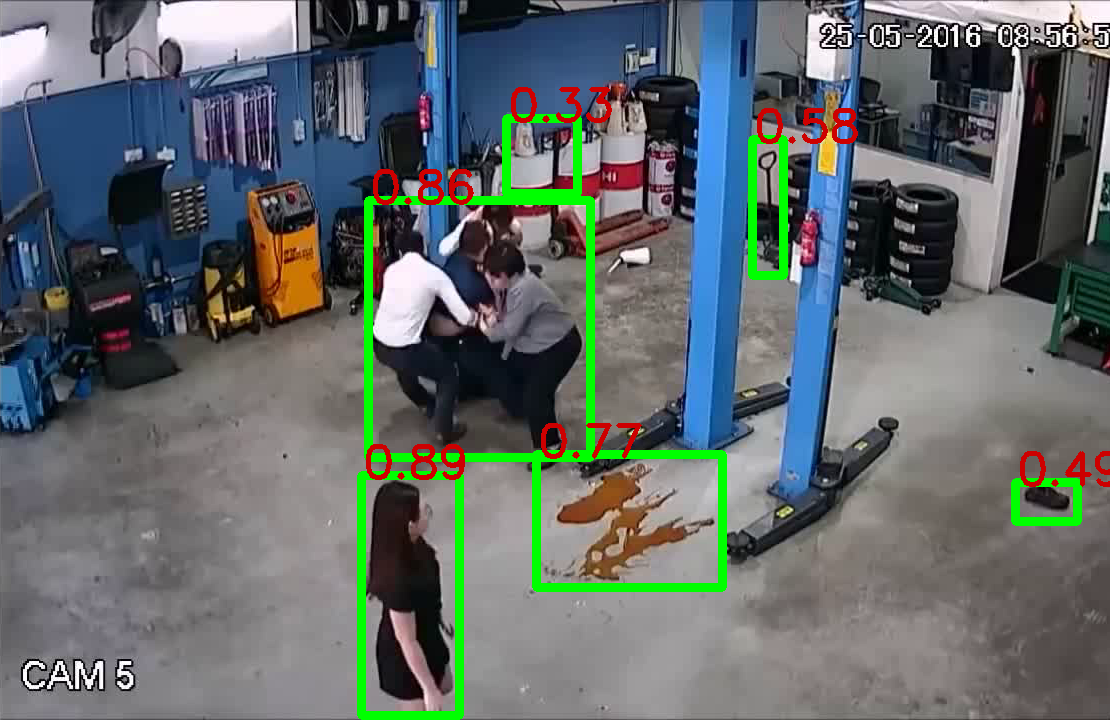}\vspace{4pt}
	   \includegraphics[width=1\linewidth]{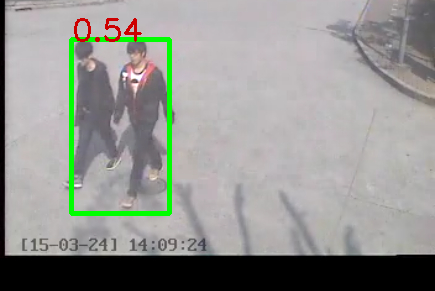}\vspace{4pt}
	   \includegraphics[width=1\linewidth]{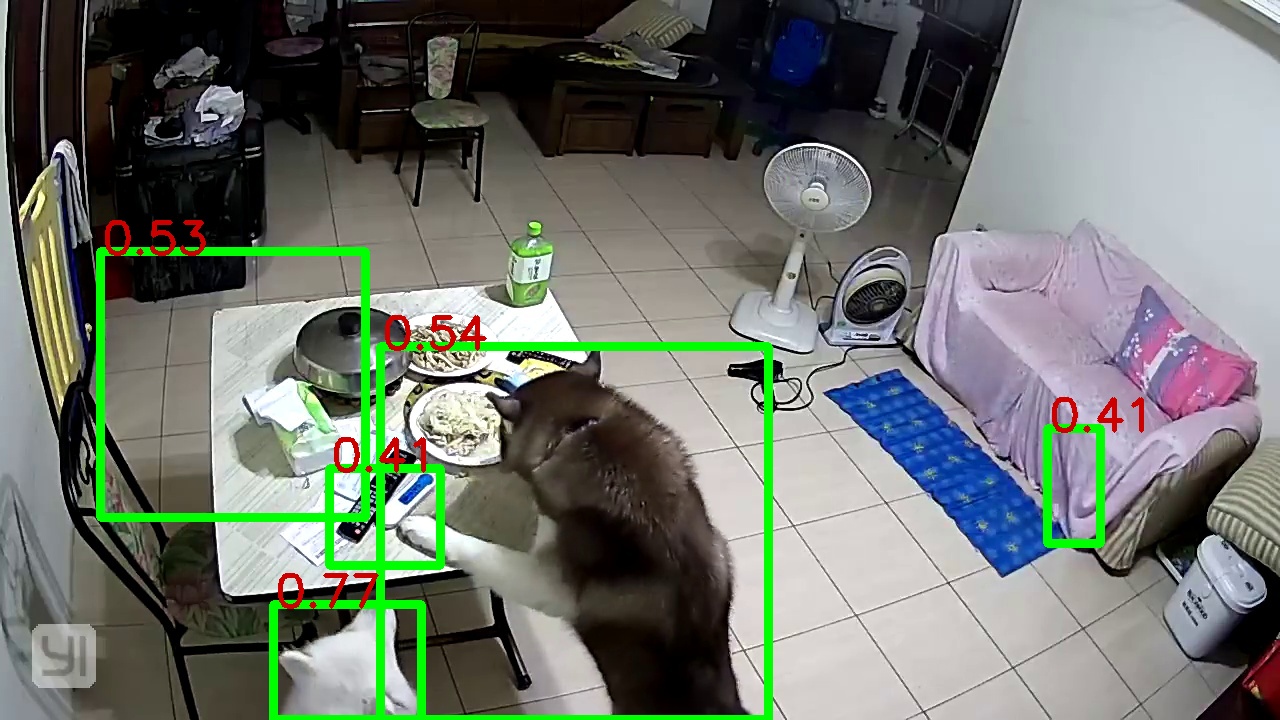}
\end{minipage}}
\label{4d}\hfill
    \subfloat[Fine-tune]{
\begin{minipage}[b]{0.12\linewidth}
\includegraphics[width=1\linewidth]{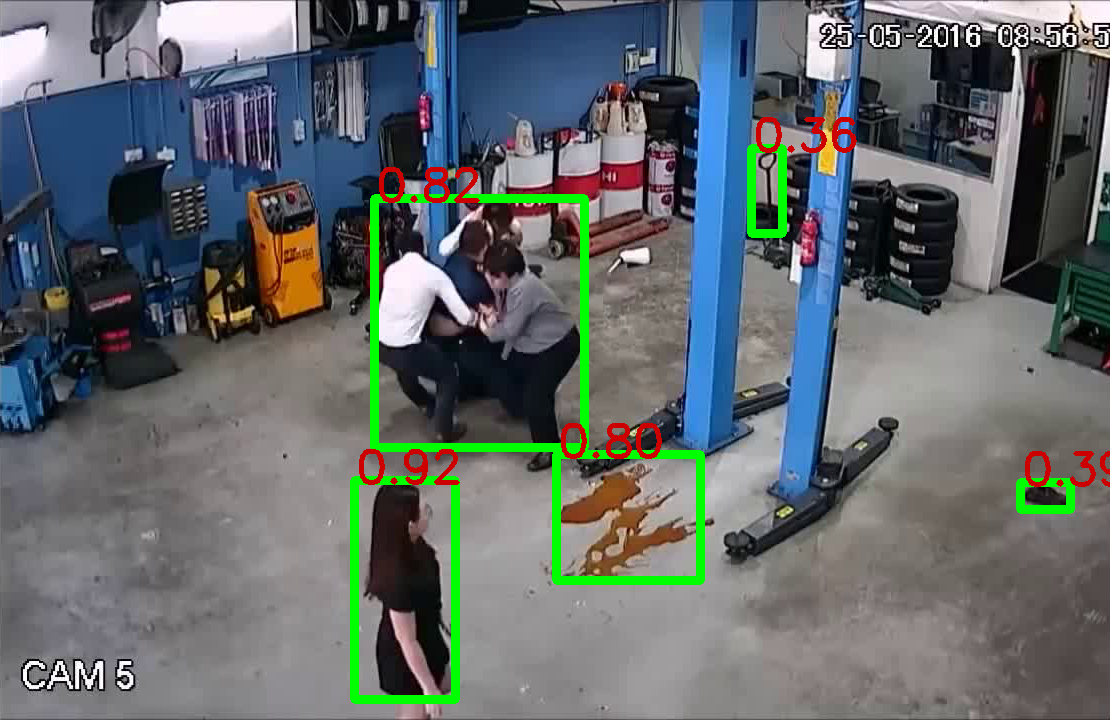}\vspace{4pt}
	   \includegraphics[width=1\linewidth]{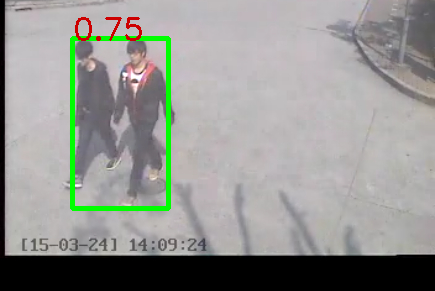}\vspace{4pt}
	   \includegraphics[width=1\linewidth]{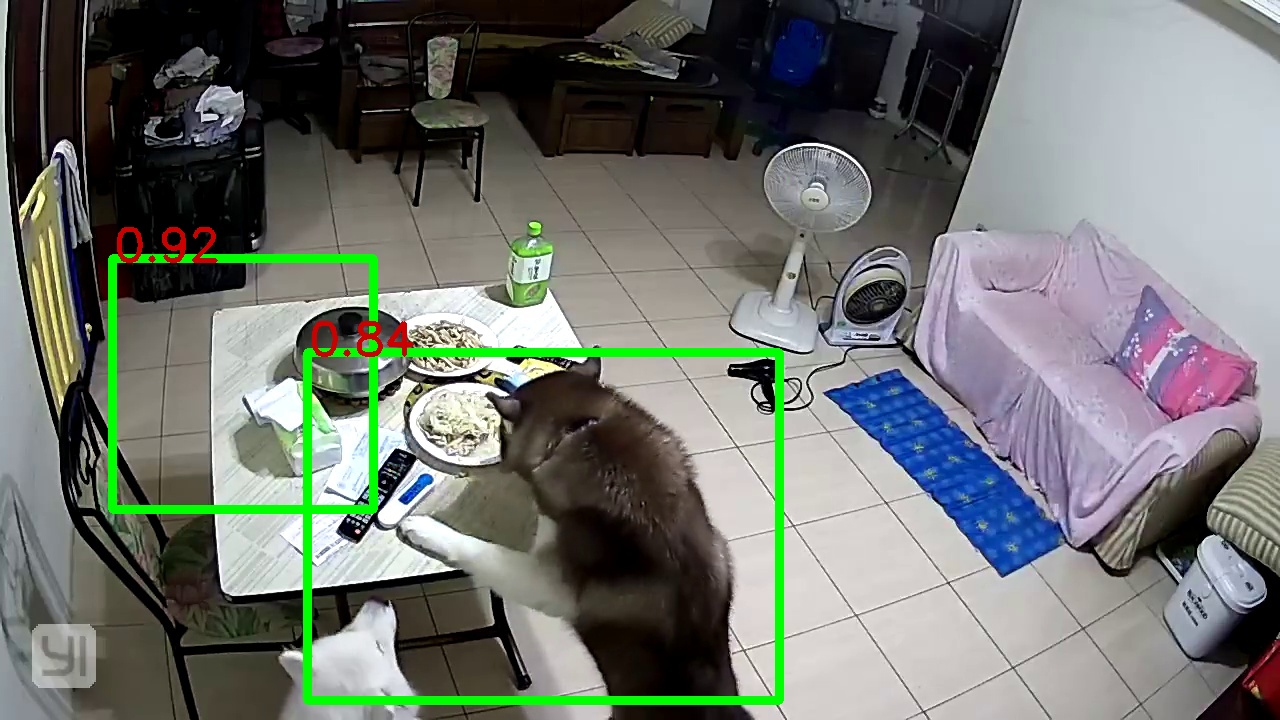}
\end{minipage}}
\label{4e}\hfill
    \subfloat[From scratch]{
\begin{minipage}[b]{0.12\linewidth}
\includegraphics[width=1\linewidth]{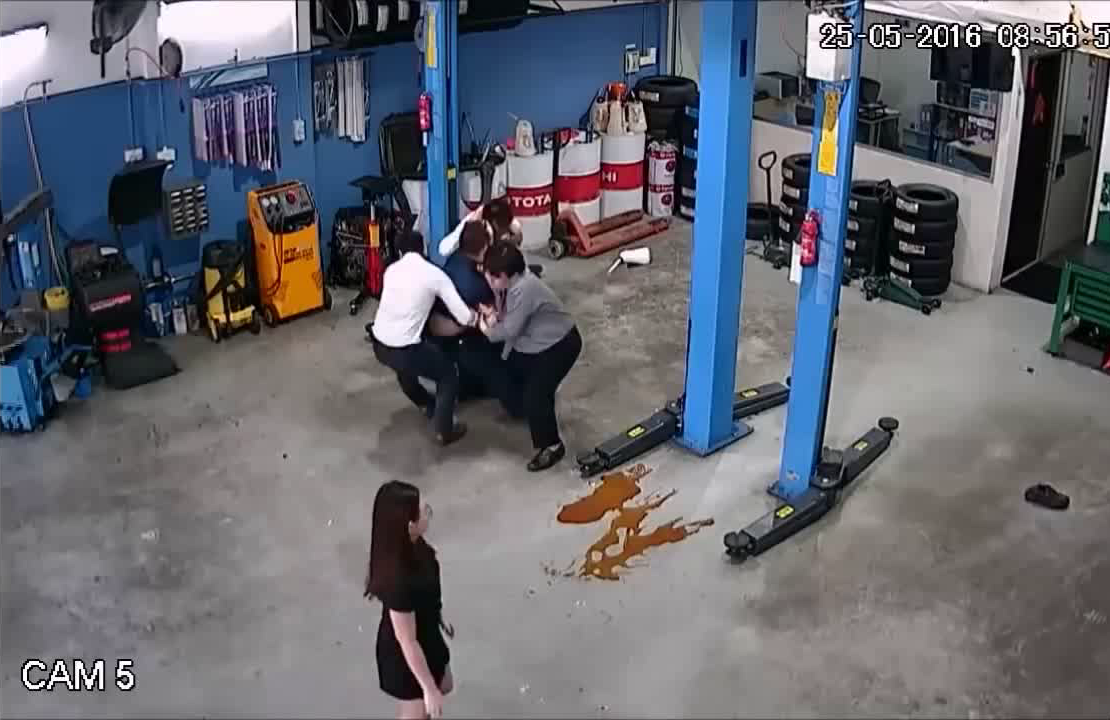}\vspace{4pt}
	   \includegraphics[width=1\linewidth]{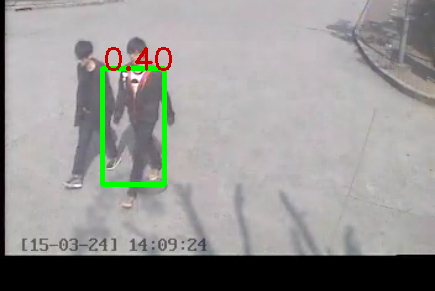}\vspace{4pt}
	   \includegraphics[width=1\linewidth]{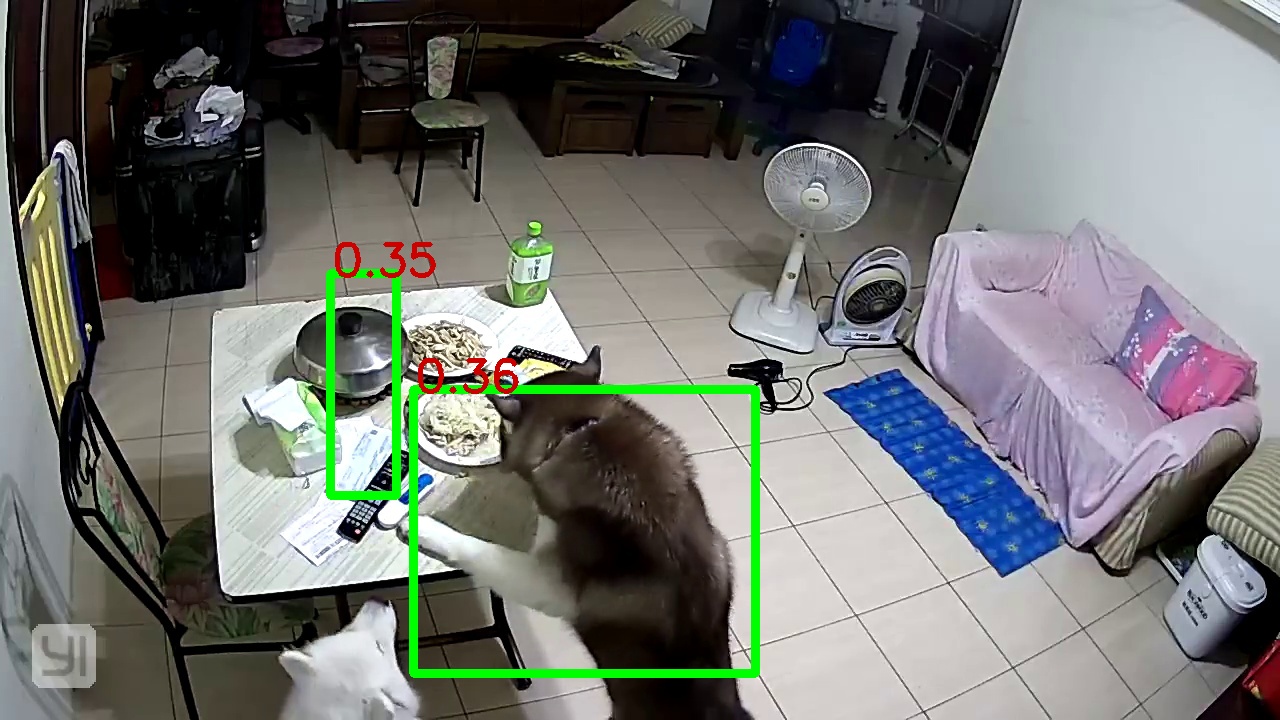}
\end{minipage}}
\label{4f}\hfill
	  \caption{Examples of detection results from different models. The top row shows a sample from Fight; the middle row shows a sample from SJ-SPID,  while the bottom is from Pets.  Images (a) are the test images, on which we use boxes to mark the differences; images (b) are the reference images; images (c) show the ground truth of image pairs; images (d) (e) (f) (g) are the detection results from the model without fine-tuning, the model with mix data, the model with fine-tuning and the model trained from scratch respectively. }
	  \label{fine-tune} 
\end{figure*} 

 \subsubsection{Utilize a few samples}
 To demonstrate that the different definitions of `change' in different scenarios become a significant obstacle to promising generalization capability, fine-tuning and mixture experiments with a few samples are designed on the three poor performance datasets Fight, SJ-SPID and Pets. For fine-tuning, $10\%$ of data regarded as a few samples from the real-world dataset are sampled to fine-tune the pretrained model from Exp.7. While for the mixture, $10\%$ of data are mixed with the synthetic data generated using Exp.7 methods and we train the model on it. To make a comparison with them, we also train models from scratch on the three samples. CUNet-EF network is selected for its better performance
 
 Figure~\ref{fine-tune} shows an example of the detection results from different models. It is obvious the model without fine-tuning confuses about `what is change' and `what is the boundary between different change'. The bounding boxes it detects are sensitive to report the noise such as shadow that we do not regard as a `change' in this scenario. While for the model trained from scratch, due to the lack of knowledge about the size of `change', the bounding boxes it detects can not cover the correct objects and even can not recognize them. As for the model with mix data and model with fine-tuning, the two methods both utilize the implicit information contained in the few samples and learn what the task focuses on to some extent.
 
 The $AP_{0.5}$ of them are shown in Table~\ref{trans}. It can be seen that these few samples can not provide enough feature information so that the model trained from scratch has a poor performance. However, when they are used with synthetic data, we can get a satisfactory result. And the fine-tuning method is more robust and better than the mixture way.

 \begin{table}[!t] \renewcommand{\arraystretch}{1.3} \caption{Experiments of a few samples on Fight, SJ-SPID and Pets} \label{trans} \centering \begin{tabular}{c|c|c|c|c} \hline \bfseries & 
  \bfseries without fine-tune &\bfseries fine-tune &\bfseries mixture &\bfseries from scratch \\ \hline
 Fight & $58.1\%$ & \cellcolor{lightgray}\textbf{$\mathbf{75.4}\%$} & $72.5\%$ &$30.2\%$ \\ 
 SJ-SPID & $71.5\%$ & \cellcolor{lightgray}\textbf{$\mathbf{95.5}\%$} & $87.0\%$ &$63.0\%$  \\   
 Pets & $35.7\%$ & \cellcolor{lightgray}\textbf{$\mathbf{80.1}\%$} & $65.6\%$ &$37.7\%$  \\ 
 \hline
\end{tabular} \end{table}

% \subsubsection{Subsubsection Heading Here}
% Subsubsection text here.

\section{Conclusion}
In this paper, we present a technique, the blend of `Regular Cropped Change with rotation and margin blur' and `Irregular Cropped Change with rotation and margin blur', to synthesize training image pairs for image change detection based on object detection. Besides, an early fusion network to tackle this task is proposed. Results show that the model trained on the synthetic data is robust on multiple testsets, while the model trained on the insufficient real-world data can not compare with it. Furthermore, a few samples experiment demonstrates that the variant definition of `change' is a hard obstacle to improving the generalization capability and utilize few samples to fine-tune the model trained on the synthetic data will overcome it and have a drastic improvement. From a practical standpoint, our technique can effectively solve the problem of insufficient training data. Besides, Nine manually labeled datasets are also provided for future research.

% conference papers do not normally have an appendix

% use section* for acknowledgment
%\section*{Acknowledgment}

%The authors would like to thank...

% trigger a \newpage just before the given reference
% number - used to balance the columns on the last page
% adjust value as needed - may need to be readjusted if
% the document is modified later
%\IEEEtriggeratref{8}
% The "triggered" command can be changed if desired:
%\IEEEtriggercmd{\enlargethispage{-5in}}

% references section

% can use a bibliography generated by BibTeX as a .bbl file
% BibTeX documentation can be easily obtained at:
% http://mirror.ctan.org/biblio/bibtex/contrib/doc/
% The IEEEtran BibTeX style support page is at:
% http://www.michaelshell.org/tex/ieeetran/bibtex/
\bibliographystyle{IEEEtran}
% argument is your BibTeX string definitions and bibliography database(s)
%\bibliography{IEEEabrv,../bib/paper}
\bibliography{ref}
%
% <OR> manually copy in the resultant .bbl file
% set second argument of \begin to the number of references
% (used to reserve space for the reference number labels box)
% \begin{thebibliography}{1}

% \bibitem{IEEEhowto:kopka}
% H.~Kopka and P.~W. Daly, \emph{A Guide to \LaTeX}, 3rd~ed.\hskip 1em plus
%   0.5em minus 0.4em\relax Harlow, England: Addison-Wesley, 1999.

% \bibitem{arXiv:Wu}
% J. Wu, Y. Ye, Y. Chen, and Z. Weng. Spot the Difference by Object Detection. 
% \emph{arXiv preprint arXiv:1801.01051 [cs.CV], 2018}

% \end{thebibliography}

% that's all folks
\end{document}